\documentclass{article}

\usepackage{amsmath}
\usepackage{amsfonts}
\usepackage{multirow}
\usepackage{graphicx}
\usepackage{caption}
\usepackage{subcaption}
\usepackage{longtable}
\usepackage{pdflscape}
\usepackage[a4paper, margin=0.7in]{geometry}
\usepackage{bm}
\usepackage{afterpage}
\usepackage{multicol}
\usepackage{float}
\usepackage{longtable}
\usepackage[toc,page]{appendix}
\usepackage{breqn}
\usepackage{lscape}
\usepackage{amssymb,mathtools}
\usepackage{rotating}
\usepackage{authblk}\usepackage{siunitx}
\usepackage[backend=bibtex, minbibnames=3, maxbibnames=99,style=numeric,citestyle=numeric-comp,sorting=none, giveninits ]{biblatex}
\DeclareNameAlias{author}{last-first}

\usepackage{verbatim}
\usepackage{xcolor}
\usepackage{hyperref}

\usepackage{titlesec}
\usepackage{enumitem}

\titleformat{\paragraph}[runin]
{\normalfont\normalsize\bfseries}{\theparagraph}{1em}{}[:]
\titlespacing*{\paragraph}
{0pt}{3.25ex plus 1ex minus .2ex}{1.5ex plus .2ex}

\providecommand{\keywords}[1]{\textbf{\textit{Keywords---}} #1}

\DeclareMathAlphabet\mathbfcal{OMS}{cmsy}{b}{n}

\title{Machine learning for detection of stenoses and aneurysms: application in a physiologically realistic virtual patient database}
\author[1]{Gareth Jones}
\author[2]{Jim Parr}
\author[1]{Perumal Nithiarasu}
\author[1, $\dagger$]{Sanjay Pant}
\affil[1]{College of Engineering, Swansea University, Swansea, United Kingdom.}
\affil[2]{McLaren Technology Centre, Woking, United Kingdom.}

\affil[$\dagger$]{Corresponding author: Sanjay.Pant@swansea.ac.uk}

\date{}      
\bibliography{article.bib}

\begin{document}

\maketitle

\begin{abstract}

\noindent This study presents an application of machine learning (ML) methods for detecting the presence of stenoses and aneurysms in the human arterial system. Four major forms of arterial disease---carotid artery stenosis (CAS), subclavian artery stenosis (SAC), peripheral arterial disease (PAD), and abdominal aortic aneurysms (AAA)---are considered. The ML methods are trained and tested on a physiologically realistic virtual patient database (VPD) containing 28,868 healthy subjects, which is adapted from the authors previous work and augmented to include the four disease forms. Six ML methods---Naive Bayes, Logistic Regression, Support Vector Machine, Multi-layer Perceptron, Random Forests, and Gradient Boosting---are compared with respect to classification accuracies and it is found that the tree-based methods of Random Forest and Gradient Boosting outperform other approaches. The performance of ML methods is quantified through the $F_1$ score and computation of sensitivities and specificities. When using all the six measurements, it is found that maximum $F_1$ scores larger than 0.9 are achieved for CAS and PAD, larger than 0.85 for SAS, and larger than 0.98 for both low- and high-severity AAAs. Corresponding sensitivities and specificities are larger than 90\% for CAS and PAD, larger than 85\% for SAS, and larger than 98\% for both low- and high-severity AAAs. When reducing the number of measurements, it is found that the performance is degraded by less than 5\% when three measurements are used, and less than 10\% when only two measurements are used for classification. For AAA, it is shown that $F_1$ scores larger than 0.85 and corresponding sensitivities and specificities larger than 85\% are achievable when using only a single measurement. The results are encouraging to pursue AAA monitoring and screening through wearable devices which can reliably measure pressure or flow-rates.\\[3pt]


\noindent \keywords{virtual patients, stenosis, aneurysm, pulse wave haemodynamics, screening, machine learning}

\end{abstract}

\section{Introduction}
\label{section_intro}

Two of the most common forms of arterial disease are stenosis, narrowing of an arterial vessel, and aneurysm, an increase in the area of a vessel. They are estimated to affect between 1\% and 20\% of the population \cite{fowkes2013comparison, shadman2004subclavian, mathiesen2001prevalence, li2013prevalence}, and ruptured abdominal aortic aneurysms alone are estimated to cause 6,000–8,000 deaths per year in the United Kingdom \cite{darwood2012twenty}. Current methods for the detection of arterial disease are primarily based on direct imaging of the vessels, which can be expensive and hence prohibitive for large-scale screening. 
If arterial disease can be detected by easily acquirable pressure and flow-rate measurements at select locations within the arterial network, then large-scale screening may be facilitated.  

It  is  likely  that  the  indicative  biomarkers of arterial disease in the pressure and flow-rate profiles consist of micro inter- and intra-measurement details.  
%
In the past, detection of arterial disease has been proposed through the analysis of waveforms in combination with mathematical models of pulse wave propagation, see for example \cite{sazonov2017novel, stergiopulos1992computer}. This, however, requires specification or identification of patient-specific network parameters, which is not easy to perform, especially at large scales.

This study explores the use of Machine Learning (ML) methods for the detection of stenoses and aneurysms in order to facilitate large scale low-cost screening/diagnosis.
A data-driven ML approach is adopted, which does not require specification of patient-specific parameters. Instead, such algorithms learn patterns and biomarkers from a labelled data set. ML has a history of being used for medical applications \cite{kononenko2001machine}. Classification algorithms have been shown to be able to predict the presence of irregularities in heart valves \cite{ccomak2007decision}, arrhythmia \cite{song2005support}, and sleep apnea \cite{khandoker2009support} from recorded time domain data. A previous study \cite{chakshu2020towards} has applied deep-learning methods to AAA classification, using a synthetic data-set created by varying seven parameters. In \cite{chakshu2020towards} accuracies of $\approx 99.9\%$ are reported for binary classification of AAA based on three pressure measurements. These studies motivate the application of ML to detect arterial disease---both stenosis and aneurysms---using only pressure and flow-rate measurements at select locations in the arterial network. A previous proof-of-concept study \cite{jones2021proof} showed promising results that ML classifiers can detect stenosis in a simple three vessel arterial network using only measurements of pressures and flow-rates. Here, these ideas are extended to a significantly larger, physiologically realistic, network of the human arterial system. All the ML methods are trained and tested on the virtual healthy  subject database proposed in \cite{jonesphysiologically2021}, which is augmented to introduce disease into the virtual subjects.


This study is organised as follows. It begins by briefly explaining the healthy VPD proposed in \cite{jonesphysiologically2021}. Modifications to this database to create four different forms of arterial disease are presented next, along with the
parameterisation of disease forms.  
This is followed by presentation of the ML methodology and metrics used for quantification of classification accuracies.
%
Finally, these accuracies are assessed when using different combinations pressure and flow-rate measurements, along with the analysis of patterns and behaviours observed in the ML classifiers.

\section{Methodology}

The ML algorithms are trained and tested on a data set containing both healthy subjects and diseased patients.

\subsection{Healthy subjects}

A physiologically realistic VPD containing healthy subjects is created in \cite{jonesphysiologically2021} and forms the starting point of this study. This database is available at \cite{jones_vpd_zenodo}. The arterial network contains 71 vessel segments and is shown in Figure 1, along with the locations where disease occurs in high prevalence, and where measurements of pressure and flow-rate can potentially be acquired \cite{jonesphysiologically2021}. The healthy patient database of \cite{jonesphysiologically2021} contains 28,868 VPs and is referred as $\text{VPD}_{\text{H}}$. Disease is introduced into these healthy arterial networks as described next.

\begin{figure}
\centering
\includegraphics[width=4.5in]{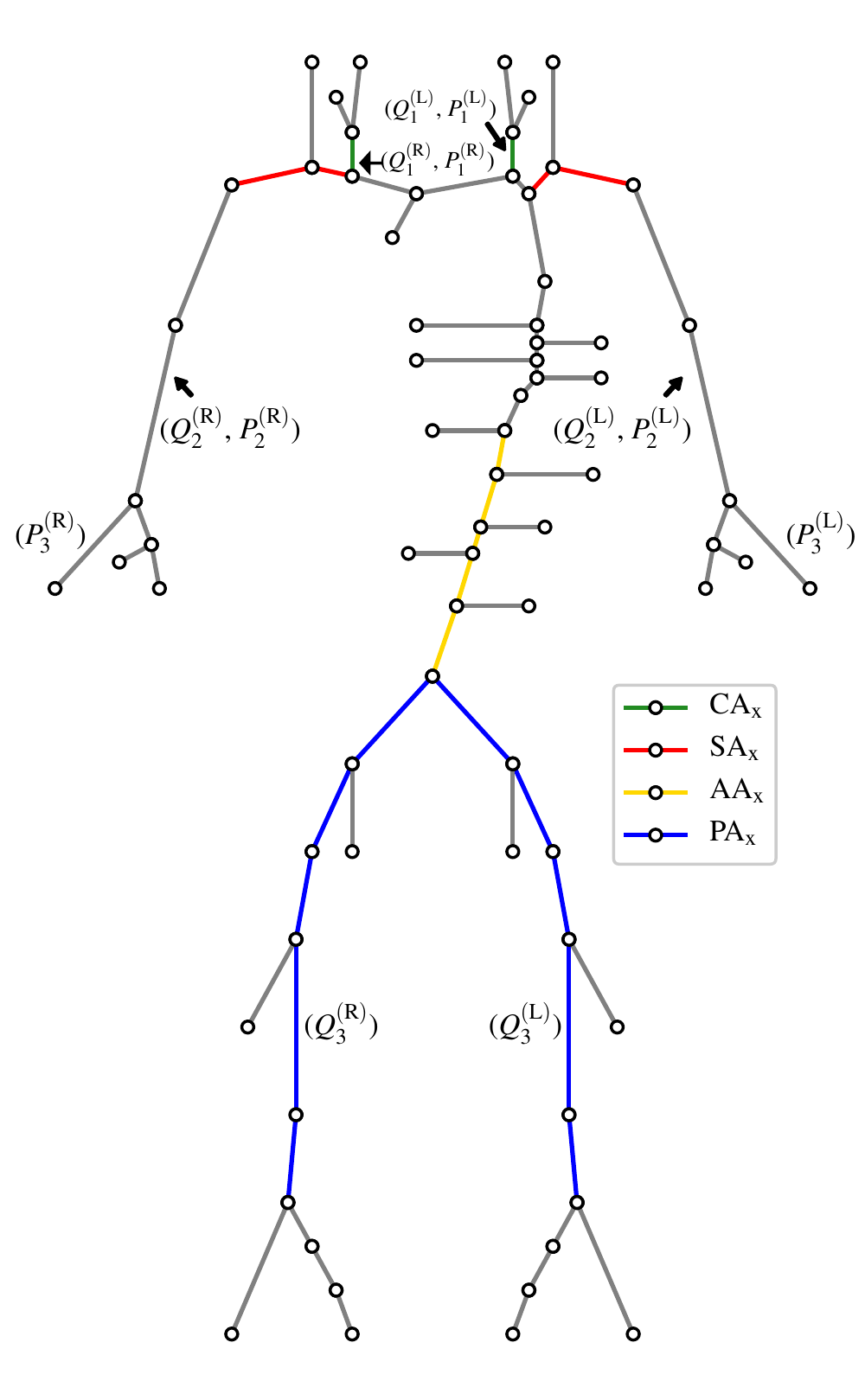}
\caption{The connectivity of the arterial network, taken from \cite{jonesphysiologically2021}. The location of the four forms of disease (see Section \ref{section_disease_forms}); and six pressure and flow-rate measurements (see section \ref{section_measurements}) are highlighted.}
\label{fig_full_network}
\end{figure}

\subsection{Creation of unhealthy VPDs}
\label{sec_unhealthy_VPD}

\subsubsection{Disease forms}
\label{section_disease_forms}

The four most common forms of arterial disease are carotid artery stenosis (CAS), subclavian artery stenosis (SAS), peripheral arterial disease (PAD, a form of stenosis), and abdominal aortic aneurysm (AAA) \cite{jonesphysiologically2021, dyken1974complete, kullo2016peripheral, aboyans2010general, chen2013disease, li2013prevalence}. Their prevalence is restricted to the following vessels and shown in Figure \ref{fig_full_network}:

\begin{itemize}[leftmargin=*]
\item \textbf{CAS} is assumed to only affect the common carotid arteries. For simplification and consistency of notation these vessels are referred to as the \textbf{carotid artery chains (CA$_{\mathbf{x}}$)}.
\item \textbf{SAS} is assumed to affect the first and second subclavian segments. These two chains of vessels (one on the right and left side) are referred to as the \textbf{subclavian artery chains (SA$_{\mathbf{x}}$)}.
\item \textbf{PAD} is assumed to affect the common iliacs; external iliacs; first and second femoral segments; and the first popliteal segments. These chains are referred to as the \textbf{peripheral artery chains (PA$_{\mathbf{x}}$)}.
\item \textbf{AAA} is assumed to affect the first to forth abdominal aorta segment. This chain of vessels is referred to as the \textbf{abdominal aortic chain (AA$_{\mathbf{x}}$)}.
\end{itemize}
%
It is assumed that each diseased VP has only one of the four forms of arterial disease. Four complementary databases corresponding to $\text{VPD}_{\text{H}}$ are constructed, each pertaining to one form of arterial disease. To create the diseased VPD corresponding to CAS, referred to as $\text{VPD}_{\text{CAS}}$, for every subject in $\text{VPD}_{\text{H}}$, disease is introduced in CA$_{\mathrm{x}}$ (i.e.~the left or right carotid artery). This is achieved by taking the arterial network 
of a subject from VPD$_{\text{H}}$, artificially introducing a stenosis  in  CA$_{\mathrm{x}}$, and then re-running the pulse-wave propagation model \cite{jonesphysiologically2021} to compute the pressure and flow-rate waveforms. Thus, $\text{VPD}_{\text{CAS}}$ contains 28,868 VPs with CAS. Similarly, the databases corresponding to SAS, PAD, and AAA are created, and referred to as $\text{VPD}_{\text{SAS}}$, $\text{VPD}_{\text{PAD}}$, and $\text{VPD}_{\text{AAA}}$, respectively. The disease severities, locations, and shapes are varied randomly across these databases as described next.

\subsubsection{Parameterisation of diseased vessels}
\label{section_disease_parameterisation}

\begin{figure}[tb]
\centering
\includegraphics[width=5in]{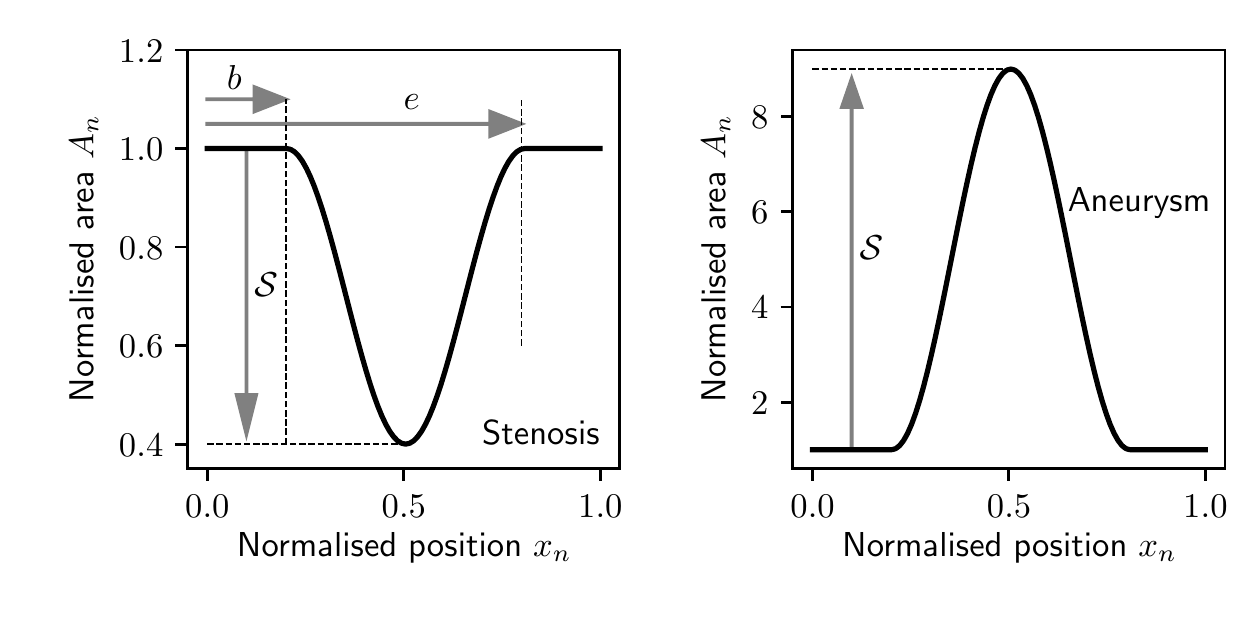}
\caption{An example of a stenosis of severity 0.6 and aneurysm of severity 8.0 are shown. These disease profiles are created with a start location of 0.2 and an end location of 0.8.}
\label{figure_area_reduction_large}
\end{figure}

The severity of stenoses (percentage reduction in area) is varied between 50\% and 95\%. The lower 50\% limit is set for the stenoses to be haemodynamically significant \cite{aboyans2010general,  subramanian2005renal} and the upper limit of 95\% reflects near total occlusion. For aneurysms, based on \cite{ernst1993abdominal} and \cite{davis2013implementation}, an allowable range of AAA severities of 4cm–6cm diameters is chosen. This corresponds to a cross sectional area variation of $12.56\text{cm}^2$--$28.27\text{cm}^2$. With the abdominal aortic area in the reference network \cite{jonesphysiologically2021} between $1.76\text{cm}^2$ and $1.09\text{cm}^2$, the corresponding AAA severities are set to vary between 713\% (12.56/1.76) and 2,593\% (28.27/1.09).
With the above ranges, parameterisation of area increase/reduction proposed in \cite{jones2021proof} is adopted, see Figure \ref{figure_area_reduction_large}. For a chain of diseased vessels (CA$_{\text{x}}$, SA$_{\text{x}}$, PA$_{\text{x}}$, or AA$_{\text{x}}$), the normalised area $A_n$ as a function of the normalised x-coordinate, $x_n$, is represented as:
\begin{equation}
A_{n}\!=\! 
\begin{cases}
\bigg(1\! \mp \! \dfrac{\mathcal{S}}{2} \bigg) \pm  \dfrac{S}{2} \cos \left(\dfrac{2 (x_n-b) \pi}{e-b}\right) & \text{for } b\leq x_n \leq e \\
\phantom{x} 1 & \text{otherwise}
\end{cases}
\label{eq_area_profile}
\end{equation}
where $\mathcal{S}$ represents the severity, $b$ represents the normalised starting location of the disease in the vessel chain, $e$ represents the normalised end location, $A_n$ is normalised with respect to the healthy version of the vessel in VPD$_{\text{H}}$, and $\pm$ creates an aneurysm or stenosis, respectively. 
In CA$_{\text{x}}$, SA$_{\text{x}}$, and PA$_{\text{x}}$, the left and right side vessels are chosen with equal probability.


\begin{figure}[tb]
\centering
\includegraphics[width=5in]{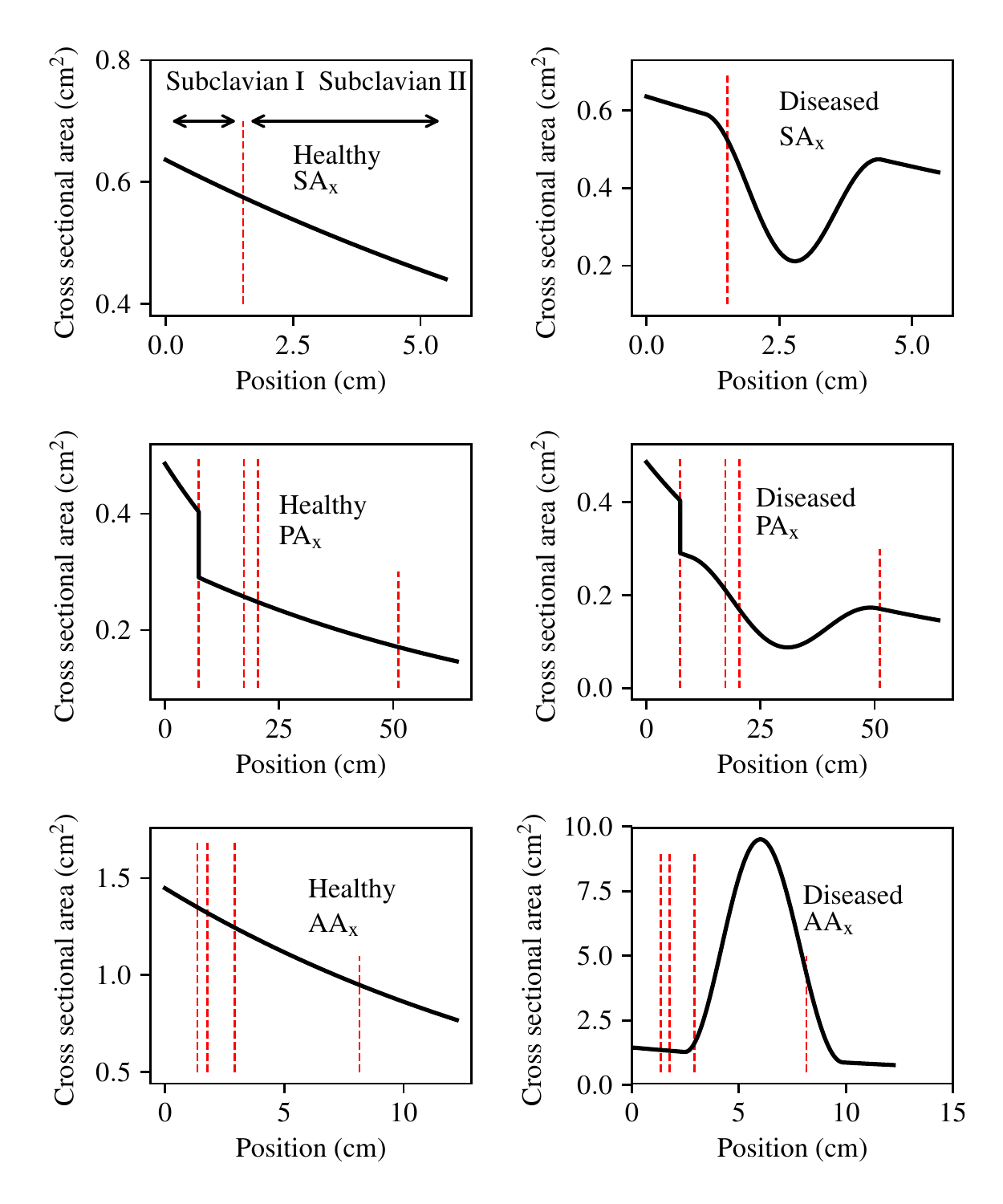}
\caption{Examples of healthy and diseased SA$_{\text{x}}$, PA$_{\text{x}}$, and AA$_{\text{x}}$ area profiles. The geometrical boundaries between vessel segments that form the chains are indicated by red dashed lines.}
\label{figure_area_profiles}
\end{figure}

 The disease severity $\mathcal{S}$, start location $b$, and end location $e$ are assigned uniform distributions based on physical considerations. To sample values for these parameters, a fourth parameter, the reference location of the disease (represented by $r$) is introduced. This is included to impose a minimum length  of 10\% of the chain length on the disease profiles. 
 Thus, the parameters for disease are sampled sequentially  from uniform distributions within the following bounds:
\begin{equation}
\text{Bounds:}
\begin{cases}
0.2 \leq r \leq 0.8,\\
0.1 \leq b \leq r-0.05, \\
r+0.05 \leq e \leq 0.9,\\[5pt]
\begin{cases} 
0.5 \leq \mathcal{S} \leq 0.95 & \text{for stenoses,}\\
7.13 \leq \mathcal{S} \leq 25.93 & \text{for aneurysms.}\\
\end{cases}
\end{cases}
\end{equation}
Based on the above parameterisation, examples of healthy and diseased SA$_{\text{x}}$, PA$_{\text{x}}$, and AA$_{\text{x}}$ area profiles are shown in the left and right columns of Figure \ref{figure_area_profiles}, respectively.

\subsection{Measurements}
\label{section_measurements}

A review of  potential measurements that can be acquired in the network is presented in \cite{jonesphysiologically2021}. Based on this, the locations at which time-varying pressure and flow-rate measurements can be acquired are shown in Figure \ref{fig_full_network} and described below. 
\begin{itemize}[leftmargin=*]
\item \textbf{Pressure in the carotid and radial arteries} measured using applanation tonometry \cite{adji2006clinical, o2015carotid}. To simplify annotation and description the right and left carotid artery pressures are referred as  $P_1^{\text{(R)}}$ and $P_1^{\text{(L)}}$, respectively. Similarly, the radial artery pressures are referred to $P_3^{\text{(R)}}$ and $P_3^{\text{(L)}}$, respectively.
\item \textbf{Pressure in the brachial arteries} estimated through reconstruction of finger arterial pressure \cite{guelen2008validation}. The right and left brachial artery pressures are referred to as $P_2^{\text{(R)}}$ and $P_2^{\text{(L)}}$ respectively.
\item \textbf{Flow-rate in the carotid, brachial, and femoral arteries} measured using Doppler ultrasound \cite{bystrom1998ultrasound, oglat2018review, radegran1997ultrasound}. The right and left carotid artery, brachial, and femoral flow-rates are referred to as $Q_1^{\text{(R)}}$, $Q_1^{\text{(L)}}$; $Q_2^{\text{(R)}}$, $Q_2^{\text{(L)}}$; and $Q_3^{\text{(R)}}$, $Q_3^{\text{(L)}}$, respectively.
\end{itemize}

\subsubsection{Provision of measurements to ML classifiers}
\label{section_measurement_provision}
Unless specified otherwise, the measurements to ML classifiers are bilateral, \textit{i.e.} when $Q_1$ is specified it is implied that both right and left carotid flow-rates are used:
\begin{equation}
Q_1 = \{Q_1^{\text{(R)}}, Q_1^{\text{(L)}}\}.
\label{equation_input_measurement}
\end{equation}
There are, therefore, a total of by six bilateral measurements available: three pressure and three flow-rates. To reduce the dimensionality required to describe each pressure or flow-rate measurement, the periodic profiles are described through a Fourier series (FS) representation:
\begin{equation}
u(t)=\sum_{n=0}^N a_n \sin (n \omega t) + b_n \cos(n \omega t),
\label{eq_FS_rep_1}
\end{equation}
where $u$ represents any pressure or flow-rate profile; $a_n$ and $b_n$ represent the $n^{\text{th}}$ sine and cosine FS coefficients, respectively; $N$ represents the truncation order; and $\omega={2 \pi}/{T}$, with $T$ as the time period of the cardiac cycle. It is found in \cite{jones2021proof} that haemodynamic profiles can be described by a FS truncated at $N=5$.  Thus, each individual measurement is described by 11 FS coefficients, and each bilateral measurement by 22 FS coefficients.


\subsection{Machine learning classifiers}

A model mapping a vector of input measurements, $\bm{x}$, to a discrete output classification, $y$, can be described as:
\begin{equation}
y = m(\bm{x}) \quad y \in  \{\mathcal{C}^{(1)}, \mathcal{C}^{(2)}\},
\label{eq_direct_model}
\end{equation}
where  $\mathcal{C}^{(j)}$ represents the $j^{\text{th}}$ possible classification. In the context of this study, the measured inputs, $\bm{x}$, represents the FS coefficients of a user defined combination of the haemodynamic measurements $\{Q_1$, $Q_2$, $Q_3$, $P_1$, $P_2$, $P_3\}$ (see Section \ref{section_measurement_provision}) taken from VPs, and the output classification represents the corresponding health of those VPs : $\mathcal{C}^{(1)}$= `healthy' and $\mathcal{C}^{(2)}$= `diseased'.  To account for large differences in magnitudes of the the components of $\bm{x}$, they are individually transformed with the Z-score standardisation method \cite{mohamad2013standardization} to have zero-mean and unit variance.

As previously stated, it assumed that in a patient disease is limited to only one of the four forms. As a first exploratory study, the ML classifiers are created for each form independently. All classifiers are therefore binary (see \cite{jones2021proof}), \textit{i.e.} four independent classifiers are trained to predict the following questions independently: \emph{``Does a VP belong to $\text{VPD}_{\text{H}}$ or $\text{VPD}_x$''}, where $x$ can be either CAS, SAS, PAD, or AAA.

\subsubsection{Training and test sets}
\label{section_train_test}

Each VP in $\text{VPD}_{\text{CAS}}$, $\text{VPD}_{\text{SAS}}$, $\text{VPD}_{\text{PAD}}$, and $\text{VPD}_{\text{AAA}}$ shares an identical underlying arterial network, apart from the diseased chain, with the corresponding healthy subject in VPD$_{\text{H}}$. It is, therefore, important to ensure that the same subset of VPs is not included in the both healthy and diseased data sets used for ML classifiers. As each form of disease is mutually exclusive, four independent training and test sets, each corresponding to one form of the disease, are constructed in the following three stages:
\begin{itemize}[leftmargin=*]
\item \textbf{Step 1:} Half of the available VPs are randomly selected from $\text{VPD}_{\text{H}}$ for inclusion within the ML data set; this is referred to as $\text{VPD}_{\text{H-ML}}$. The unhealthy VPs corresponding to the remaining unused half are taken from the appropriate unhealthy VPD ($\text{VPD}_{\text{CAS}}$, $\text{VPD}_{\text{SAS}}$, $\text{VPD}_{\text{PAD}}$, or $\text{VPD}_{\text{AAA}}$) and incorporated into the ML data set. These data sets are referred to as $\text{VPD}_{\text{CAS-ML}}$, $\text{VPD}_{\text{SAS-ML}}$, $\text{VPD}_{\text{PAD-ML}}$, or $\text{VPD}_{\text{AAA-ML}}$.
\item \textbf{Step 2:} The data sets of Step 1 are combined to create four complete data sets each containing 50\% healthy and 50\%, unhealthy VPs:
\begin{enumerate}
\item $\text{VPD}_{\text{H-ML}}\cup \text{VPD}_{\text{CAS-ML}}$
\item $\text{VPD}_{\text{H-ML}}\cup \text{VPD}_{\text{SAS-ML}}$
\item $\text{VPD}_{\text{H-ML}}\cup \text{VPD}_{\text{PAD-ML}}$
\item $\text{VPD}_{\text{H-ML}}\cup \text{VPD}_{\text{AAA-ML}}$
\end{enumerate}
\item \textbf{Step 3:} The four data sets of Step 2 are randomly split into a training set containing 2/3 of all the VPs in the data set, and a test set containing 1/3 of all the VPs.
\end{itemize} 
The performance of all ML classifiers  is evaluated using a five fold validation. For each fold, the same data set from Step 2 is used but different subsets are sampled in Step 3 for training and testing. 

\subsubsection{ML methods}

Six different ML methods are employed. These six methods are random forest, gradient boosting, 
naive Bayes' , support vector machine, logistic regression , and multi-layer perceptron. Note that the last of these, the multi-layer perceptron, may be considered as a deep learning method. These methods are chosen as they encompass a range of probabilistic and non-probabilistic applications of different modelling approaches, see Table \ref{table_classifier_characteristics}, while requiring minimal problem specific optimisation. Since standard versions and implementations of these methods are employed without any modifications, methodological details of these methods are not presented in this study. Instead the reader is referred to the following references for methodological details:
\begin{enumerate}
\item Random Forest (RF) \cite{liaw2002classification, breiman2001random}
\item Gradient Boosting (GB) \cite{friedman2001greedy, elith2008working}
\item Naive Bayes' (NB) \cite{rish2001empirical, rish2001analysis}
\item Support Vector Machine (SVM) \cite{kecman2005support}
\item Logistic Regression (LR) \cite{sperandei2014understanding, hilbe2009logistic, jones2021proof}
\item Multi-layer Perceptron (MLP) \cite{murtagh1991multilayer}
\end{enumerate}
All implementations of the above algorithms in the Python package Scikits-learn \cite{pedregosa2011scikit} are used. Some of these methods require optimisation of the hyper-parameters. This is described after presenting performance quantification metrics in the next section.

\begin{table*}
\begin{center}
\def\arraystretch{1.2}
\begin{tabular}{| c | c |c |}
\hline
\textbf{Modelling approach} & \textbf{Non-probabilistic} & \textbf{Probabilistic}\\
\hline
\textbf{Tree-based} & RF & GB \\
\hline
\textbf{Kernel-based} & SVM &  \\
\hline
\textbf{Bayesian} &  & NB\\
\hline 
\textbf{Neuron-based} &  &  LR, MLP\\
\hline 
\end{tabular}
\caption{The four different modelling approaches and how each classification method aligns with these approaches.}
\label{table_classifier_characteristics}
\end{center}
\end{table*}

\subsubsection{Quantification of results}

\begin{figure}[tb]
\centering
\includegraphics[width=2.5in]{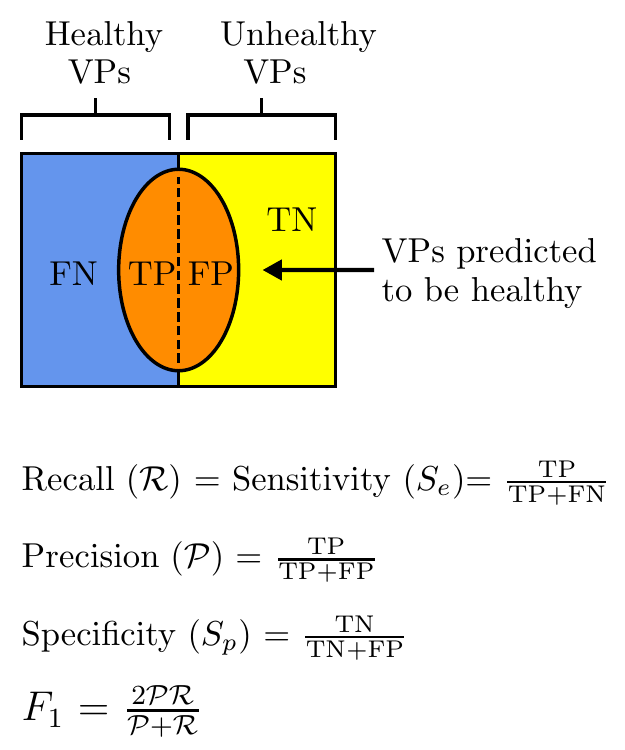}
\caption{The relationship between sensitivity, specificity, recall, and precision. TP: True Positive, representing VPs belonging to a classification correctly identified; FN: False Negative, representing VPs belonging to a classification incorrectly identified: FP: False positive, representing VPs not belonging to a classification incorrectly identified; and TN: True Negative, representing VPs not belonging to a classification correctly identified.}
\label{figure_precision_recall}
\end{figure}

Classifier performance is assessed by two metrics:  \emph{sensitivity} and \emph{specificity} in combination; and the $F_1$ score. Figure \ref{figure_precision_recall} shows the definition of {sensitivity}, {specificity}, and $F_1$ score, along with the related concepts of \emph{precision} and \emph{recall} commonly used in the assessment of classifiers. It is desirable to have both sensitivities and specificities to be high. Similarly, a higher $F_1$ score is desirable. Since the $F_1$ score is a single scalar metric that balances both precision and recall, it is a good metric to compare classifiers when tuning the hyper-parameters of ML algorithms. For a discussion on these metrics and their relevance, please refer to \cite{jones2021proof}.

\subsection{Hyperparameter optimisation}

The architecture of LR, NB, and SVM classifiers can all be considered to be problem independent. While these three algorithms are able to undergo varying levels of problem specific optimisation, the underlying structure of the classifier usually does not change. The architectures of RF, MLP, and GB classifiers, however, are dependent on the specific problem. The architecture choices for the classifiers and associated hyper-parameter optimisation is described next.

\subsubsection{LR, SVM, and NB} 
For LR, the `LIBLINEAR' solver offered by the Scikits-learn \cite{pedregosa2011scikit} package is chosen.
In the case of SVM, a kernel is typically chosen to map the input measurements to a higher order feature space \cite{jakkula2006tutorial}. All SVM classifiers use a radial basis function kernel \cite{scholkopf1997comparing}. In the case of NB, the distribution of input measurements across the data set is chosen to be Normal \cite{murphy2006naive}. 

\subsubsection{Random Forest}
In the case of RF, the number of trees in the ensemble and the maximum depth of each tree is optimised. To optimise these two hyper-parameters, a  grid search is carried out. A grid is constructed by discretising the possible number of trees within the ensemble between 10 and 400 at intervals of 10; and the possible depth of each tree between 20 and 200 at intervals of 10. RF classifiers are trained for every combination with all six pressure and flow-rate measurements (see Section \ref{section_measurement_provision}) across all the four forms of arterial disease. The hyper-parameters describing the architecture that produces the highest $F_1$ score is  found for each form of the disease, and this combination of hyper-parameters is then chosen for all subsequent classifiers. 
The optimal hyper-parameters for each of the four forms of disease are shown in Table \ref{table_RF_gridsearch}, along with the $F_1$ score achieved by each.

It is unlikely that a single architecture will consistently produce the best results when varying the combination of input measurements. In this study, re-optimisation of the hyper-parameters when varying input measurement combination is avoided to minimise computational cost. It should be noted, however, that further improvements in classification accuracy may be possible with such re-optimisation. 

\begin{table}[htbp]
\begin{center}
\def\arraystretch{1.2}
\begin{tabular}{| c | c c |c |}
\hline
\textbf{Disease} & \textbf{Trees} & \textbf{Depth} & \textbf{$\bm{F}_1$}\\
\hline
CAS & 100 & 80 & 0.8878\\
SAS & 150 & 80 &  0.8292\\
PAD & 100 & 100 & 0.8935 \\
AAA & 100 & 50 & 0.9912\\

\hline 
\end{tabular}
\caption{The hyper-parameters describing the architecture of the RF classifiers that produce the highest $F_1$ scores, when using all six pressure and flow-rate measurements. 
}
\label{table_RF_gridsearch}
\end{center}
\end{table} 

\subsubsection{Gradient Boosting}

Similar to RF architecture, the GB architecture is optimised for the problem of this study by varying the number of trees within the ensemble and the maximum depth of each tree. A grid search is carried out to find the the combination producing the highest $F_1$ score when using all the six input measurements.  It is common for GB classifiers to use  weaker, shallower decision trees (relative to RF classifiers) to deliberately create high bias and low variance \cite{hastie2009elements}. The possible depth of each tree is, therefore, discretised between 2 and 20 at intervals of 1. As a high number of trees is not required to compensate for over fitting, contrary to the RF method, the possible number of trees within the ensemble is discretised between 10 and 100 at intervals of 10. The optimal hyper-parameters for each of the four forms of disease are shown in Table \ref{table_GB_gridsearch}.

\begin{table}[htbp]
\begin{center}
\def\arraystretch{1.2}
\begin{tabular}{| c | c c |c |}
\hline
\textbf{Disease} & \textbf{Trees} & \textbf{Depth} & \textbf{$\bm{F}_1$}\\
%
\hline
CAS & 100 & 6 & 0.9343\\
SAS & 100 & 7 &  0.8574\\
PAD & 100 & 10 & 0.9187 \\
AAA & 80 & 7 & 0.9970\\

\hline 
\end{tabular}
\caption{The hyper-parameters describing the architecture of the GB classifiers that produce the highest $F_1$ scores, when using all six pressure and flow-rate measurements. 
}
\label{table_GB_gridsearch}
\end{center}
\end{table}

\subsubsection{Multi-layer perceptron}

In the case of MLP, the number of neurons within each hidden layer, and the number of hidden layers is optimised to create the optimal architecture for the classification problem of this study. For simplicity, it is assumed that all the hidden layers contain an identical number of neurons. Similar to RF and GB, the hyper-parameters that produce the highest $F_1$ score are found through a grid search. The number of neurons within each layer is discretised between 10 and 200 at intervals of 10, and the number of hidden layers is discretised between 1 and 6 at intervals of 1. The optimal hyper-parameters found for each of the four forms of disease are shown in Table \ref{table_MLP_gridsearch}. It shows that relative to RF and GB, there is less consistency in the maximum $F_1$ scores achieved by MLP--- it classifies AAA and CAS to high levels of accuracies, but  performs relatively poorly for SAS and PAD. 

\begin{table}[htbp]
\begin{center}
\def\arraystretch{1.2}
\begin{tabular}{| c | c c |c |}
\hline
\textbf{Disease} & \textbf{Neurons} & \textbf{Depth} & \textbf{$\bm{F}_1$}\\
\hline

CAS & 60 & 4 & 0.7785\\
SAS & 190 & 2 &  0.6040\\
PAD & 120 & 2 & 0.6681 \\
AAA & 30 & 2 & 0.9785\\

\hline 
\end{tabular}
\caption{The hyper-parameters describing the architecture of the MLP classifiers that produce the highest $F_1$ scores, when using all six pressure and flow-rate measurements. 
}
\label{table_MLP_gridsearch}
\end{center}
\end{table}

\section{Results and discussion}
\label{sec_results_and_discussion}

There are 63 possible combinations of input measurements that can be provided to a ML classifier from the six bilateral pressure and flow-rate measurements (see Section \ref{section_measurement_provision}). A combination search is performed for each of the four forms of disease. For every combination of input measurements all the six ML classification methods are trained, and then subsequently tested to quantify their performance. The average $F_1$ score, sensitivity, and specificity for each case  across five folds are recorded. Combinations of interest are then further analysed.

\begin{sidewaysfigure*}
\centering
\vspace{1cm}
\includegraphics[width=9in]{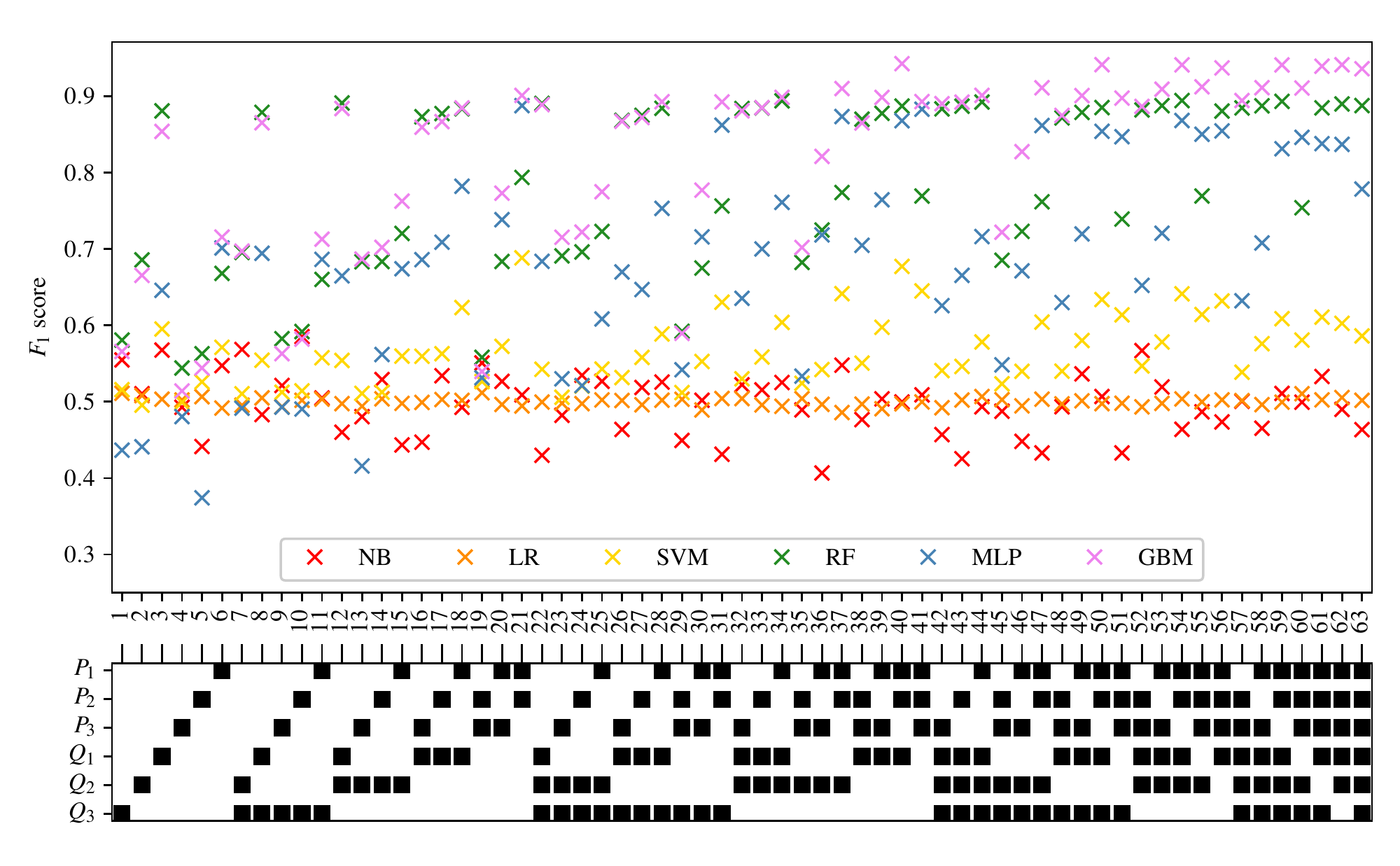}
\caption{The $F_1$ scores achieved for CAS using each combination of bilateral input measurements are shown. Measurements included within each combination are highlighted with a black square.}
\label{fig_combination_inputs_carotid}
\end{sidewaysfigure*}

\begin{sidewaysfigure*}
\centering
\vspace{1cm}
\includegraphics[width=9in]{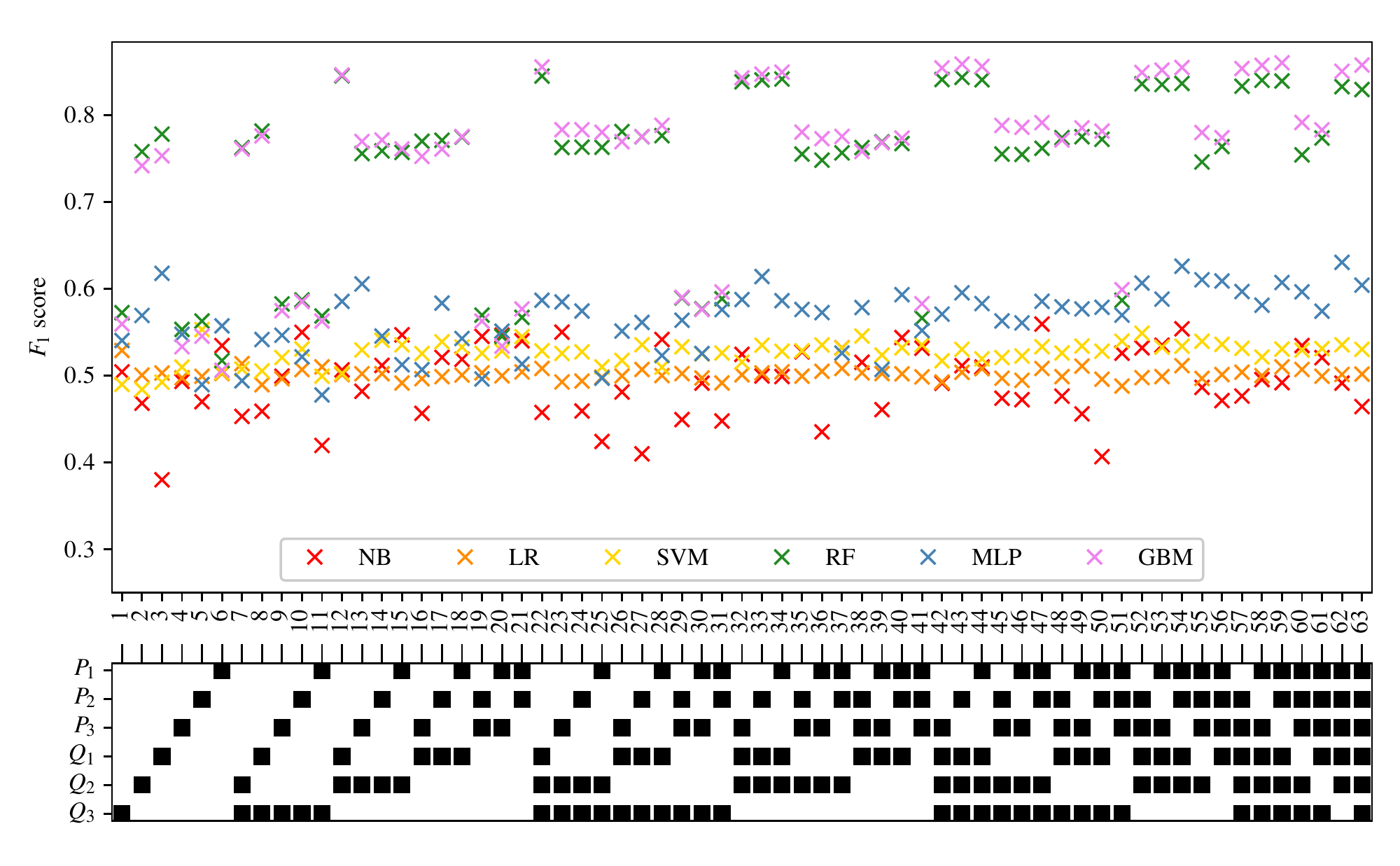}
\caption{The $F_1$ scores achieved for SAS using each combination of bilateral input measurements are shown. Measurements included within each combination are highlighted with a black square.}
\label{fig_combination_inputs_SS}
\end{sidewaysfigure*}

\begin{sidewaysfigure*}
\centering
\vspace{1cm}
\includegraphics[width=9in]{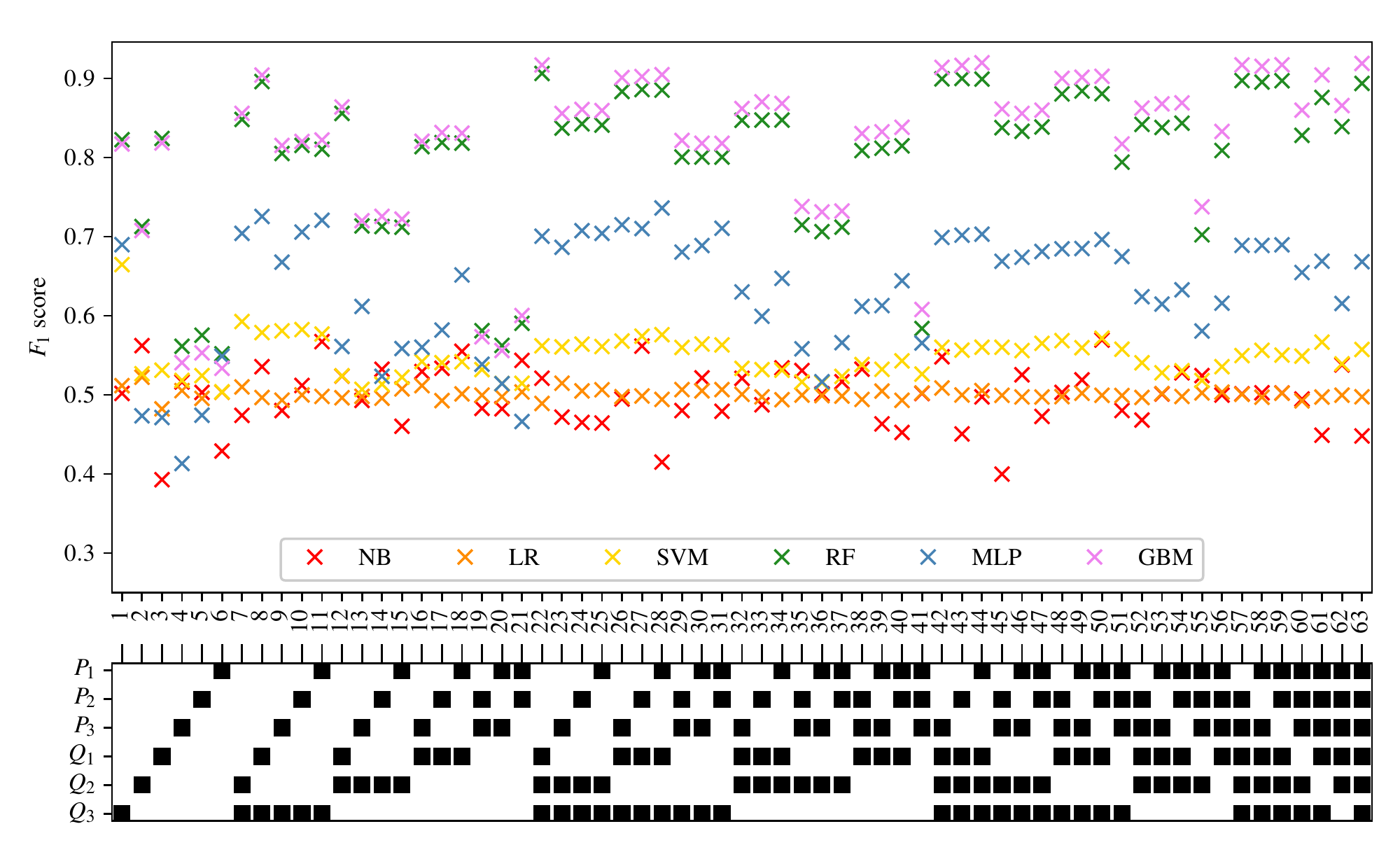}
\caption{The $F_1$ scores achieved for PAD using each combination of bilateral input measurements are shown. Measurements included within each combination are highlighted with a black square.}
\label{fig_combination_inputs_PAD}
\end{sidewaysfigure*}

\begin{sidewaysfigure*}
\centering
\vspace{1cm}
\includegraphics[width=9in]{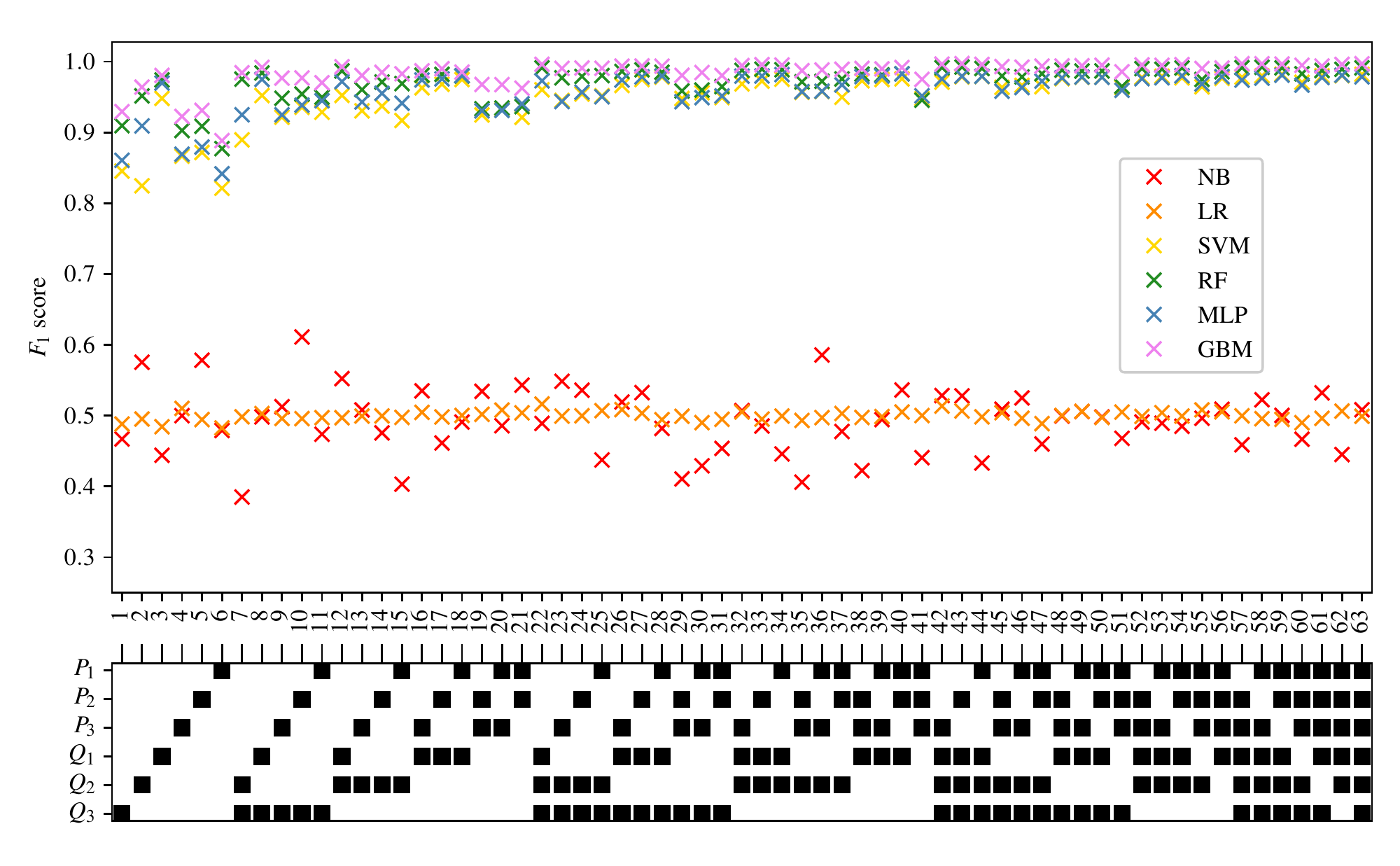}
\caption{The $F_1$ scores achieved for AAA using each combination of bilateral input measurements are shown. Measurements included within each combination are highlighted with a black square.}
\label{fig_combination_inputs_AAA}
\end{sidewaysfigure*}

The full tables of results achieved for CAS, SAS, PAD, and AAA classification are shown in Appendices \ref{appendix_CS}, \ref{appendix_SS}, \ref{appendix_PAD}, and \ref{appendix_AAA} respectively. The $F_1$ score achieved by each ML method and combination of input measurements are visually shown for CAS, SAS, PAD, and AAA classification in Figures \ref{fig_combination_inputs_carotid}, \ref{fig_combination_inputs_SS}, \ref{fig_combination_inputs_PAD}, and \ref{fig_combination_inputs_AAA} respectively.
%
They show that for all forms of arterial disease, NB and LR classifiers consistently produce low accuracy. It has previously been shown in the PoC  \cite{jones2021proof} that the partition between the pressure and flow-rate profiles taken from healthy and stenosed patients is likely to be non-linear. The fact that LR consistently produces low accuracy results supports this finding, as LR is the only linear classification method used. The finding that NB classifiers also produce low accuracy classification is also consistent with the results of the PoC \cite{jones2021proof}, which found that the NB method is poorly suited to the problem of distinguishing between hemodynamic profiles.
On the contrary, across all the four forms of disease, the tree based methods (RF and GB) consistently produce high accuracy results. This finding is in contradiction to the finding in the PoC \cite{jones2021proof}, and is likely due to the inadequate architecture optimisation or because of the unsuitability of RF on a smaller network used in the PoC \cite{jones2021proof}. 
The fact that both RF and GB classifiers are producing high accuracy classification in this study suggests that not only are tree based methods well suited to distinguishing between haemodynamic profiles, but also emphasises the importance of adequate architecture optimisation.

There is less consistency in the results achieved by SVM and MLP classifiers when detecting different forms of disease. SVM classifiers produce accuracies comparable with RF and GB classifiers in the case of AAA detection, however low accuracy results for the three other forms of disease. MLP classifiers produce accuracies comparable with RF and GB classifiers in the case of CAS and AAA detection, however relatively low accuracy results for SS and PAD classification. 
Overall, it is found that tree-based methods of RF and GB perform best, with GB performance slightly superior to that of RF.

\subsection{Measurement combinations}

To investigate the importance of both the number of input measurements provided to the ML algorithms and the specific combination of measurements, the average $F_1$ scores achieved by all classifiers when providing only one, two, three, four, five, or six input measurements are found. In each case, the specific combinations that achieve the maximum and minimum $F_1$ scores are also recorded. These results for different forms of disease are presented next.

\subsubsection{CAS classification}
\label{sec_carotid_num_inputs}

\begin{figure}[htbp]
\centering
\includegraphics[width=4in]{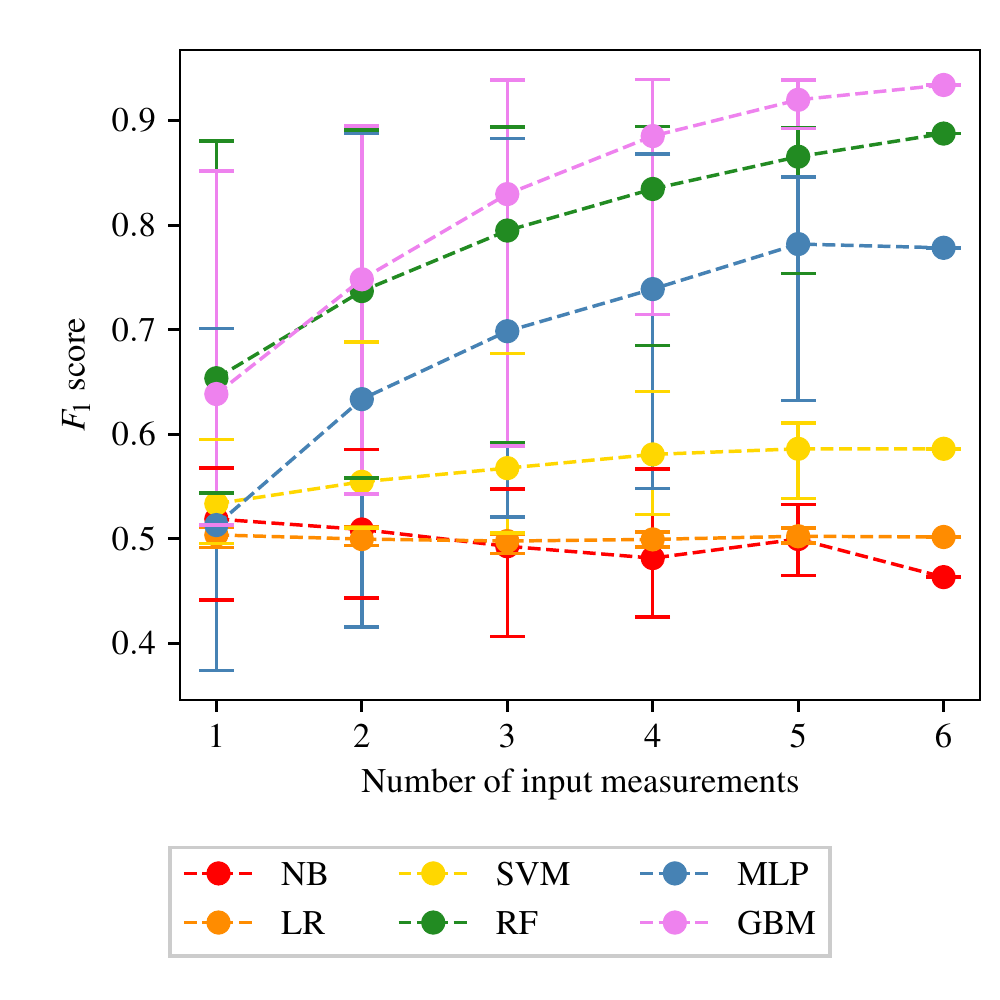}
\caption{The average, maximum, and minimum $F_1$ score achieved by all classifiers trained using different numbers of input measurements are shown for carotid artery stenosis classification. The central markers represent the average score achieved, while the error bars indicate the upper and lower limits.}
\label{fig_number_measurements_carotid}
\end{figure}

The average, maximum and minimum $F_1$ score achieved when providing different number of input measurements for CAS classification are shown in Figure \ref{fig_number_measurements_carotid}. 
It shows that NB, LR, and SVM classifiers consistently produce an $F_1$ score of approximately 0.5, which is comparable to naive classification, \textit{i.e.} randomly assigning the health of VPs with an equal probability to each outcome. SVM performs slightly better with $F_1$ scores averaging 0.5 -- 0.6. The other three classification methods (RF, MLP, and GB) perform significantly better with $F_1$ scores generally averaging between 0.7 and 0.95 and showing a clear increase in the average $F_1$ score as the number of input measurements increases. While the average and minimum $F_1$ score achieved by RF and GB classifiers continuously increases, the maximum $F_1$ score achieved can be seen to quickly reach a plateau (at one input measurement for RF, and three input measurements for GB). For a fixed number of measurements, the wide range of $F_1$ scores in Figure \ref{fig_number_measurements_carotid} across all classifiers suggests that specific combinations of measurements may be more important than others for optimal classification. To explore this further, the combinations of input measurements that produce the highest $F_1$ scores and the corresponding accuracies when employing the RF and GB methods are shown in Table \ref{table_max_CAS}. Two observations are made from this table. First that for a fixed number of measurements, the best combinations are not identical for the two methods. For example, when two measurements are used the best combination for RF is ($Q_2$, $Q_1$) while the best combination for GB is ($P_2$, $P_1$). This suggests that the best combination of measurements is likely dependent on the particular ML method chosen. Second, some patterns stand out with respect to which measurements may be more informative than others. For example, across the Table \ref{table_max_CAS} $Q_1$ appears in 11 out of 12 combinations and $P_1$ appears in 8 out of 12 combinations. This suggests that $Q_1$ is most informative about identifying the presence of CAS followed by $P_1$.  Physiologically, this is not surprising as $Q_1$ and $P_1$ are flow-rates and pressures in the carotid arteries and the disease under consideration is carotid artery stenosis. It is encouraging that the ML methods are indeed placing more importance to the relevant physiological measurements.
In fact, it is remarkable that RF and GB both achieve $F_1$ scores above 0.85 and sensitivities and specificities larger than 85\% with only this measurement. Also notable is that these accuracies can be taken to beyond 93\% (see GB row for 3 measurements in Table \ref{table_max_CAS}) when adding 2 more measurements as long as the additional two measurements are carefully chosen.
%
%

\begin{table}[htbp]
\begin{center}
\def\arraystretch{1.2}
\begin{tabular}{ |c | c c |c  c c |}
\hline
\textbf{Number of input } & \textbf{Method} & \textbf{Combination} & \textbf{$F_1$} & \textbf{Sens.} &  \textbf{Spec. }\\
\textbf{measurements }&  &  &\textbf{score} &  \textbf{}  &  \textbf{} \\
\hline
\multirow{2}{*}{\textbf{1}} & RF & ($Q_1$) & 0.8809 & 0.8704 & 0.8893\\
 & GB & ($Q_1$) & 0.8521 & 0.8547 & 0.8502\\
\multirow{2}{*}{\textbf{2}} & RF & ($Q_2$, $Q_1$) & 0.8913 & 0.8765 & 0.9032 \\
 & GB & ($P_2$, $P_1$) & 0.8950 & 0.9026 & 0.8889 \\
\multirow{2}{*}{\textbf{3}} & RF & ($Q_2$, $Q_1$, $P_1$) & 0.8941 & 0.8825 & 0.9035\\
 & GB & ($Q_1$, $P_2$, $P_1$) & 0.9389 & 0.9433 & 0.9351\\
\multirow{2}{*}{\textbf{4}} & RF & ($Q_2$, $Q_1$, $P_2$, $P_1$) & 0.8944 & 0.8858 & 0.9015 \\
 & GB & ($Q_3$, $Q_1$, $P_2$, $P_1$) & 0.9395 & 0.9417 & 0.9376 \\
\multirow{2}{*}{\textbf{5}} & RF & ($Q_3$, $Q_2$, $Q_1$, $P_2$, $P_1$) & 0.8934 & 0.8858 & 0.8996\\
 & GB & ($Q_2$, $Q_1$, $P_3$, $P_2$, $P_1$) & 0.9391 & 0.9416 & 0.9370 \\
\multirow{2}{*}{\textbf{6}} & RF & \multirow{2}{*}{($Q_3$, $Q_2$, $Q_1$, $P_3$, $P_2$, $P_1$)} & 0.8878 & 0.8747 & 0.8984\\
& GB &  & 0.9343 & 0.9364 & 0.9325\\
\hline 
\end{tabular}
\caption{The combinations of input measurements that produce the maximum $F_1$ scores when providing one to six input measurements and employing the RF and GB methods to detect CAS. The corresponding sensitivities and specificities are also included.}
\label{table_max_CAS}
\end{center}
\end{table}

An interesting pattern to note is that while the average and minimum $F_1$ score achieved by MLP classifiers continuously increases in Figure \ref{fig_number_measurements_carotid}, the maximum $F_1$ score decreases beyond three input measurements. The maximum $F_1$ scores achieved by MLP classifiers, and the corresponding sensitivities and specificities, when using three to six input measurements are shown in Table \ref{table_max_min_CS_MLP}. It shows that the decrease in $F_1$ scores is also accompanied by an associated decrease in both the sensitivities and specificities, as opposed to the balance between them (increase in sensitivity and decrease in specificity and vice versa).
This behaviour is unusual as intuitively more input measurements should generally provide more information. This finding may suggest that MLP classifiers are able to extract maximum information from the haemodynamic profiles when using as little as three input measurements, and may be susceptible to over fitting when using more than three measurements, thereby leading to less generalisation capabilities and consequently decreased accuracies.

\begin{table}[htbp]
\begin{center}
\def\arraystretch{1.2}
\begin{tabular}{ |c | c |c  c c |}
\hline
\textbf{Number of input } & \textbf{Combination} & \textbf{$F_1$} & \textbf{Sensitivity} &  \textbf{Specificity }\\
\textbf{measurements }&   &\textbf{score} &  \textbf{}  &  \textbf{} \\
\hline
\textbf{3} & ($P_3$, $P_2$, $P_1$) & 0.8831 & 0.8731 & 0.8911\\
\textbf{4} &  ($Q_3$, $Q_1$, $P_2$, $P_1$) & 0.8683 & 0.8538 & 0.8545 \\
\textbf{5} & ($Q_3$, $Q_2$, $P_3$, $P_2$, $P_1$) & 0.8463 & 0.8308 & 0.8577\\
\textbf{6} & ($Q_3$, $Q_2$, $Q_1$, $P_3$, $P_2$, $P_1$) & 0.7785 & 0.7916 & 0.7703\\
\hline 
\end{tabular}
\caption{The combinations of input measurements that produce the maximum $F_1$ scores when providing three to six input measurements and employing the MLP method to detect CAS. The corresponding sensitivities and specificities are also included.}
\label{table_max_min_CS_MLP}
\end{center}
\end{table}

\subsubsection{SAS classification}
\label{sec_SS_num_inputs}

\begin{figure}[htbp]
\centering
\includegraphics[width=4in]{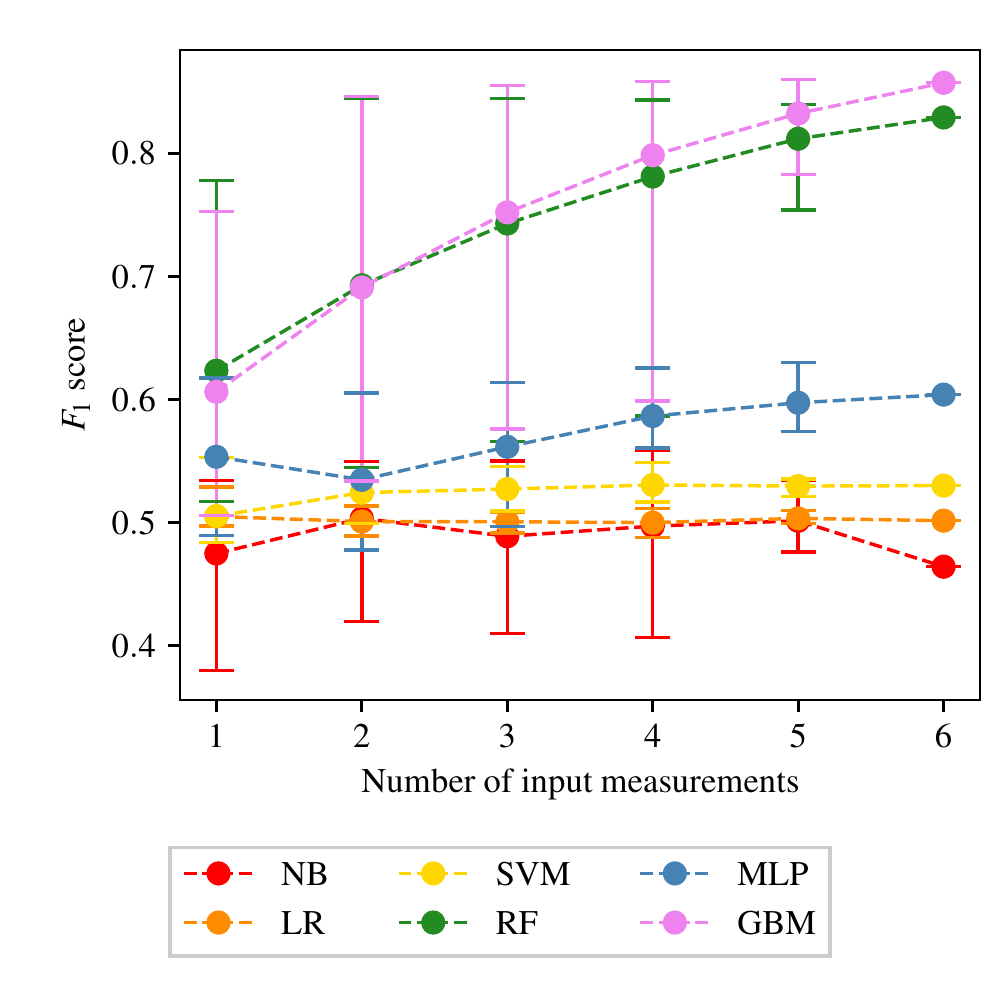}
\caption{The average, maximum, and minimum $F_1$ score achieved by all classifiers trained using different numbers of input measurements are shown for SAS classification. The central markers represent the average score achieved, while the error bars indicate the upper and lower limits.} 
\label{fig_number_measurements_SS}
\end{figure}

The results of the analysis for SAS classification are shown in Figure \ref{fig_number_measurements_SS}. As is seen in the case of CAS classification, Figure \ref{fig_number_measurements_SS} shows that NB, LR, and SVM classifiers consistently produce accuracies comparable to naive classification, irrespective of the number of input measurements used. A clear difference between Figures \ref{fig_number_measurements_carotid} and \ref{fig_number_measurements_SS} is the accuracy achieved by MLP classifiers. Compared to the CAS case, the MLP performance is further degraded for SAS, while still being better than NB, LR, and SVM, although only marginally.
%

A high degree of similarity can be seen between the behaviours of RF and GB classifiers for CAS and SAS. Figure \ref{fig_number_measurements_SS} shows that the average and minimum $F_1$ score achieved by RF and GB classifiers continuously increases as the number of input measurements used increases. The maximum $F_1$ score achieved is seen to quickly reach an asymptotic limit (at three input measurements for both RF and GB classifiers). Peak $F_1$ score of approximately 0.85 is achieved by GB, along with sensitivities and specificities higher than 85\%.

The combination of input measurements that produce the highest $F_1$ scores and the corresponding accuracies are shown in Table \ref{table_max_SAS}. It shows a higher degree of consistency between the best combinations for the two methods relative to the case for CAS, i.e. the best combinations are generally identical (or with minimal differences) between RF and GB.
It also shows that $Q_1$ is particularly informative, with this measurement appearing in all of the best combinations. Physiologically this may be due to its proximity to the disease location.

\begin{table}[htbp]
\begin{center}
\def\arraystretch{1.2}
\begin{tabular}{ |c | c c |c  c c |}
\hline
\textbf{Number of input } & \textbf{Method} & \textbf{Combination} & \textbf{$F_1$} & \textbf{Sens.} &  \textbf{Spec. }\\
\textbf{measurements }&  &  &\textbf{score} &  \textbf{}  &  \textbf{} \\
\hline
\multirow{2}{*}{\textbf{1}} & RF & ($Q_1$) & 0.7779 & 0.7582 & 0.7905\\
 & GB & ($Q_1$) & 0.7529 & 0.7224 & 0.7714\\
\multirow{2}{*}{\textbf{2}} & RF & ($Q_2$, $Q_1$) & 0.8450 & 0.8374 &  0.8507\\
 & GB & ($Q_2$, $Q_1$) & 0.8461 & 0.8293 & 0.8585 \\
\multirow{2}{*}{\textbf{3}} & RF & ($Q_3$, $Q_2$, $Q_1$) & 0.8447 & 0.8271 & 0.8576\\
 & GB & ($Q_3$, $Q_2$, $Q_1$) & 0.8552 & 0.8453 & 0.8626\\
\multirow{2}{*}{\textbf{4}} & RF & ($Q_3$, $Q_2$, $Q_1$, $P_2$) & 0.8432 & 0.8303 & 0.8527 \\
 & GB & ($Q_3$, $Q_2$, $Q_1$, $P_2$) & 0.8585 & 0.8487 & 0.8660 \\
\multirow{2}{*}{\textbf{5}} & RF & ($Q_3$, $Q_2$, $Q_1$, $P_3$, $P_1$) & 0.8399 & 0.8256 & 0.8504\\
 & GB & ($Q_3$, $Q_2$, $Q_1$, $P_2$, $P_1$) & 0.8600 & 0.8525 &  0.8657\\
\multirow{2}{*}{\textbf{6}} & RF & \multirow{2}{*}{($Q3$, $Q_2$, $Q_1$, $P_3$, $P_2$, $P_1$)} & 0.8292 & 0.8102 & 0.8427\\
& GB & & 0.8574 & 0.8504 & 0.8627\\
\hline 
\end{tabular}
\caption{The combinations of input measurements that produce the maximum $F_1$ scores when providing one to six input measurements and employing the RF and GB methods to detect SAS. The corresponding sensitivities and specificities are also included.}
\label{table_max_SAS}
\end{center}
\end{table}

\subsubsection{PAD classification}
\label{sec_PAD_num_inputs}

\begin{figure}[htbp]
\centering
\includegraphics[width=4in]{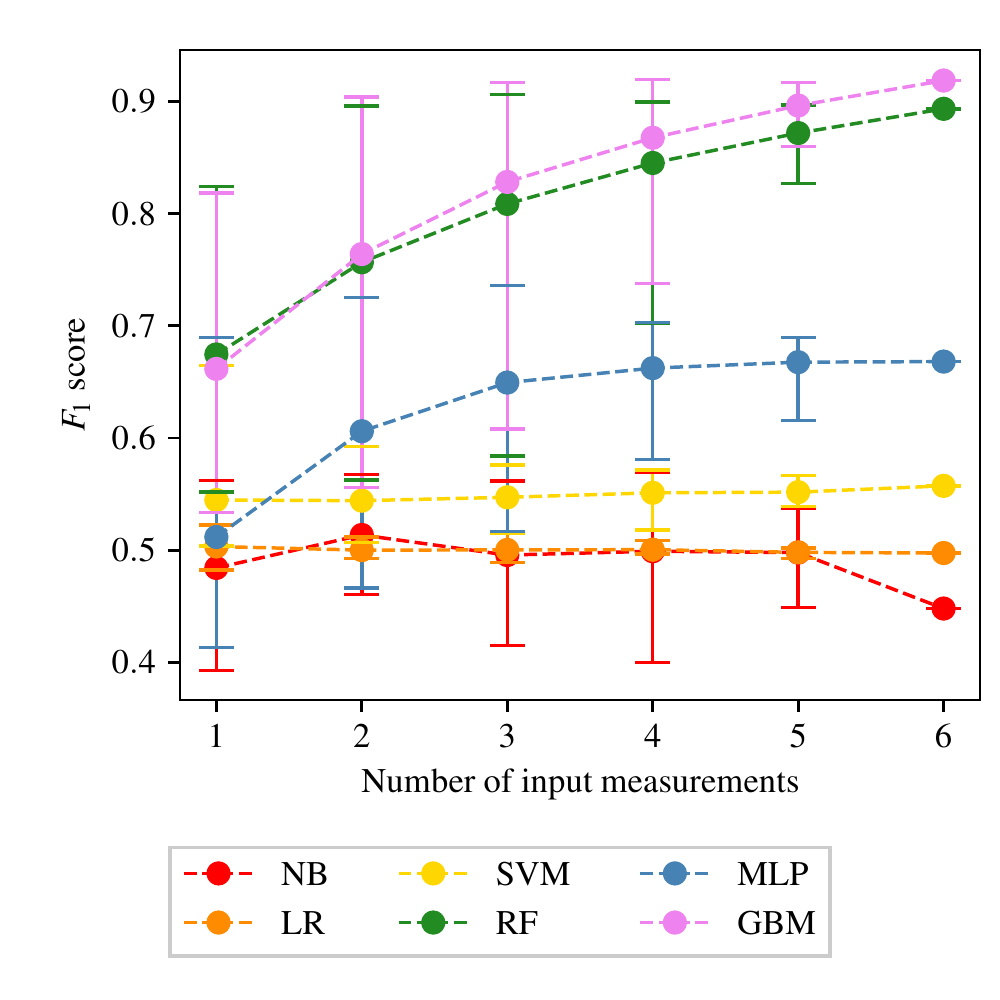}
\caption{The average, maximum, and minimum $F_1$ score achieved by all classifiers trained using different numbers of input measurements are shown for PAD classification. The central markers represent the average score achieved, while the error bars indicate the upper and lower limits.}
\label{fig_number_measurements_PAD}
\end{figure}

The results for PAD classification are shown in Figure \ref{fig_number_measurements_PAD}. Comparing Figures \ref{fig_number_measurements_SS} and \ref{fig_number_measurements_PAD}, a high degree of similarity is seen between the behaviours of SAS and PAD classification. As is previously seen for SAS classification, Figure \ref{fig_number_measurements_PAD} shows that the NB, LR, and SVM methods are all consistently producing accuracies comparable to naive classification. While the MLP method performs slightly better than the naive classification, the accuracy still remains relatively low. High accuracy can be seen in Figure \ref{fig_number_measurements_PAD} for the two tree based methods of RF and GB. As has been previously seen for CAS and SAS, while the average and minimum $F_1$ score achieved by the RF and GB methods increases as the number of input measurements increases, the maximum $F_1$ score achieved quickly reaches an asymptotic limit (at 3 input measurements for both the RF and GB methods).

The combination of input measurements that produce the highest $F_1$ scores for PAD classification when employing the RF and GB methods are shown in Table \ref{table_max_PAD}. Table \ref{table_max_PAD} not only shows good consistency between the combinations of input measurements that produce the highest $F_1$ scores when employing each of the two ML methods, but also good agreement with the combinations presented in Table \ref{table_max_SAS}. Very similar combinations of input measurements (with some minor differences) can be seen to produce the highest $F_1$ score when providing all numbers of input measurements. As has previously been observed in Tables \ref{table_max_CAS} and \ref{table_max_SAS}, the input measurement $Q_1$ appears to be most informative, appearing in all the best scoring classifiers. Since the location of $Q_1$ is far from the location of disease, it is not obvious why this measurement is particularly informative of PAD.

\begin{table}[htbp]
\begin{center}
\def\arraystretch{1.2}
\begin{tabular}{ |c | c c |c  c c |}
\hline
\textbf{Number of input } & \textbf{Method} & \textbf{Combination} & \textbf{$F_1$} & \textbf{Sens.} &  \textbf{Spec. }\\
\textbf{measurements }&  &  &\textbf{score} &  \textbf{}  &  \textbf{} \\
\hline
\multirow{2}{*}{\textbf{1}} & RF & ($Q_1$) & 0.8240 & 0.8959 & 0.8320\\
 & GB & ($Q_1$) & 0.8183 & 0.8126 & 0.8214\\
\multirow{2}{*}{\textbf{2}} & RF & ($Q_3$, $Q_1$) & 0.8140 & 0.8825 & 0.9068 \\
 & GB & ($Q_3$, $Q_1$) & 0.9041 & 0.8950 & 0.9117 \\
\multirow{2}{*}{\textbf{3}} & RF & ($Q_3$, $Q_2$, $Q_1$) & 0.9061 & 0.8885 & 0.9208\\
 & GB & ($Q_3$, $Q_2$, $Q_1$) & 0.9168 & 0.9055 & 0.9265\\
\multirow{2}{*}{\textbf{4}} & RF & ($Q_3$, $Q_2$, $Q_1$, $P_2$) & 0.8997 & 0.8868 &  0.9104\\
 & GB & ($Q_3$, $Q_2$, $Q_1$, $P_1$) & 0.9196 & 0.9068 & 0.9306 \\
\multirow{2}{*}{\textbf{5}} & RF & ($Q_3$, $Q_2$, $Q_1$, $P_3$, $P_2$) & 0.8971 & 0.8802 & 0.9110\\
 & GB & ($Q_3$, $Q_2$, $Q_1$, $P_2$, $P_1$) & 0.9170 & 0.9041 &  0.9281\\
\multirow{2}{*}{\textbf{6}} & RF & \multirow{2}{*}{($Q3$, $Q_2$, $Q_1$, $P_3$, $P_2$, $P_1$)} & 0.8935 & 0.8813 & 0.9035\\
& GB & & 0.9187 & 0.9102 & 0.9261\\
\hline 
\end{tabular}
\caption{The combinations of input measurements that produce the maximum $F_1$ scores when providing one to six input measurements and employing the RF and GB methods to detect PAD. The corresponding sensitivities and specificities are also included.}
\label{table_max_PAD}
\end{center}
\end{table}

\subsubsection{AAA classification}
\label{sec_AAA_num_inputs}

\begin{figure}[htbp]
\centering
\includegraphics[width=4in]{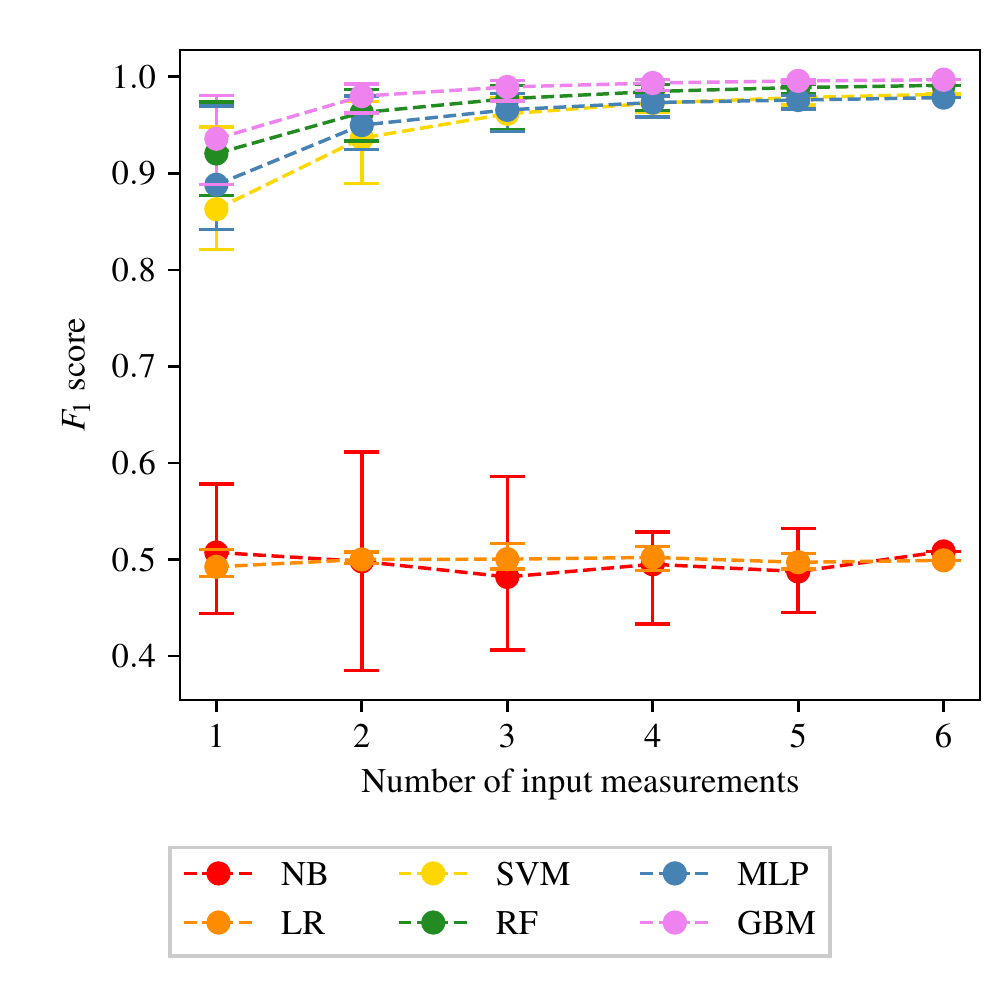}
\caption{The average, maximum, and minimum $F_1$ score achieved by all classifiers trained using different numbers of input measurements are shown for AAA classification. The central markers represent the average score achieved, while the error bars indicate the upper and lower limits.}
\label{fig_number_measurements_AAA}
\end{figure}

The results for AAA classification are shown in Figure \ref{fig_number_measurements_AAA}. As has been previously seen for all of the three other forms of disease the NB, and LR classifiers consistently produce accuracies comparable to naive classification, irrespective of the number of input measurements used. The consistency of this finding (as seen in Figures \ref{fig_number_measurements_carotid}, \ref{fig_number_measurements_SS}, and \ref{fig_number_measurements_PAD}), irrespective of the form of disease being classified, highlights both the importance of non-linear partitions between healthy and unhealthy VPs and the unsuitability of the NB method for distinction between haemodynamic profiles.\\

In the case of AAA classification the SVM, RF, MLP, and GB methods consistently produce good accuracies. Figure \ref{fig_number_measurements_AAA} shows that these methods produce high accuracies even with a single input measurement. While there is some increase in the average $F_1$ score as the number of input measurements increases, due to the very high initial average $F_1$ score achieved (when using a single input measurement) this increase is limited (as the $F_1$ score can not exceed 1).
Two possible reasons of the higher accuracies in aneurysm classification relative to stenosis classification are:
\begin{itemize}[leftmargin=*]
\item Aneurysms, owing to an increase in area as opposed to decrease in the area for stenoses, may actually produce more significant or consistent biomarkers in the pressure and flow-rate profiles. This hypothesis is supported by \cite{low2012improved}, which found that even low severity AAAs have a global impact on the pressure and flow-rate profiles. 
\item While the severities of aneurysms cannot be directly compared to severities of stenosis, it may be that the severity of aneurysms in $\text{VPD}_{\text{AAA}}$ are disproportionately large relative to the severities of stenoses. The significance of any indicative biomarkers introduced into pressure and flow-rate profiles is likely to be proportional to the severity of the change in area.
This implies that the increase in vessel area of 712\%--2,593\% in $\text{VPD}_{\text{AAA}}$ is perhaps on the extreme end of aneurysm severity, thereby making the classifications relatively easier. This is further explored in section \ref{sec_low_sev_AAA}.
\end{itemize}
%
The combination of input measurements that produce the highest $F_1$ scores when providing one to six input measurements and employing the RF and GB methods are shown for AAA classification in Table \ref{table_max_AAA}. It shows that $F_1$ scores range from 0.97--0.997 and sensitivities and specificities range from 99\% to 99.8\%. Due to the high accuracies across all the number of measurements, the analysis of specific combinations is not very meaningful. However, the measurement $Q_1$ again appears in all the best combinations. It should also be noted that the  high accuracies for AAA classification are also consistent with those reported in \cite{chakshu2020towards}, where deep-learning methods are applied on a VPD created by varying seven network parameters, and classification accuracies of $\approx 99.9\%$ are reported.\\

Overall, the results show that the physiological changes to the waveforms induced by both stenosis and aneurysms \cite{stergiopulos1992computer,low2012improved} are well captured by the data-driven machine learning methods.


\begin{table}[tb]
\begin{center}
\def\arraystretch{1.2}
\begin{tabular}{ |c | c c |c  c c |}
\hline
\textbf{Number of input } & \textbf{Method} & \textbf{Combination} & \textbf{$F_1$} & \textbf{Sens.} &  \textbf{Spec. }\\
\textbf{measurements }&  &  &\textbf{score} &  \textbf{}  &  \textbf{} \\
\hline
\multirow{2}{*}{\textbf{1}} & RF & ($Q_1$) & 0.9741 & 0.9654 & 0.9825\\
 & GB & ($Q_1$) & 0.9805 & 0.9799 & 0.9811\\
\multirow{2}{*}{\textbf{2}} & RF & ($Q_2$, $Q_1$) & 0.9868 & 0.9810 & 0.9926 \\
 & GB & ($Q_2$, $Q_1$) & 0.9928 & 0.9919 & 0.9938 \\
\multirow{2}{*}{\textbf{3}} & RF & ($Q_3$, $Q_2$, $Q_1$) & 0.9912 & 0.9864 & 0.9961\\
 & GB & ($Q_3$, $Q_2$, $Q_1$) & 0.9962 & 0.9954 & 0.9970\\
\multirow{2}{*}{\textbf{4}} & RF & ($Q_3$, $Q_2$, $Q_1$, $P_2$) & 0.9923 & 0.9879 & 0.9967 \\
 & GB & ($Q_3$, $Q_2$, $Q_1$, $P_2$) & 0.9972 & 0.9959 & 0.9986 \\
\multirow{3}{*}{\textbf{5}} & RF & ($Q_3$, $Q_2$, $Q_1$, $P_3$, $P_1$) & 0.9920 & 0.9873 & 0.9967\\
 & \multirow{2}{*}{GB} & ($Q_3$, $Q_2$, $Q_1$, $P_3$, $P_2$) & \multirow{2}{*}{0.9970} & 0.9959 & 0.9981 \\
  &  & ($Q_3$, $Q_2$, $Q_1$, $P_3$, $P_1$) &  & 0.9963 & 0.9978 \\
\multirow{2}{*}{\textbf{6}} & RF & \multirow{2}{*}{($Q3$, $Q_2$, $Q_1$, $P_3$, $P_2$, $P_1$)} & 0.9912 & 0.9861 & 0.9964\\
& GB & & 0.9970 & 0.9959 & 0.9981\\
\hline 
\end{tabular}
\caption{The combinations of input measurements that produce the maximum $F_1$ scores when providing one to six input measurements and employing the RF and GB methods to detect AAA. The corresponding sensitivities and specificities are also included.}
\label{table_max_AAA}
\end{center}
\end{table}

\subsection{Importance of carotid artery flow-rate}

\label{section_carotid_flow}

\begin{figure}[tb]
\centering
\includegraphics[width=4in]{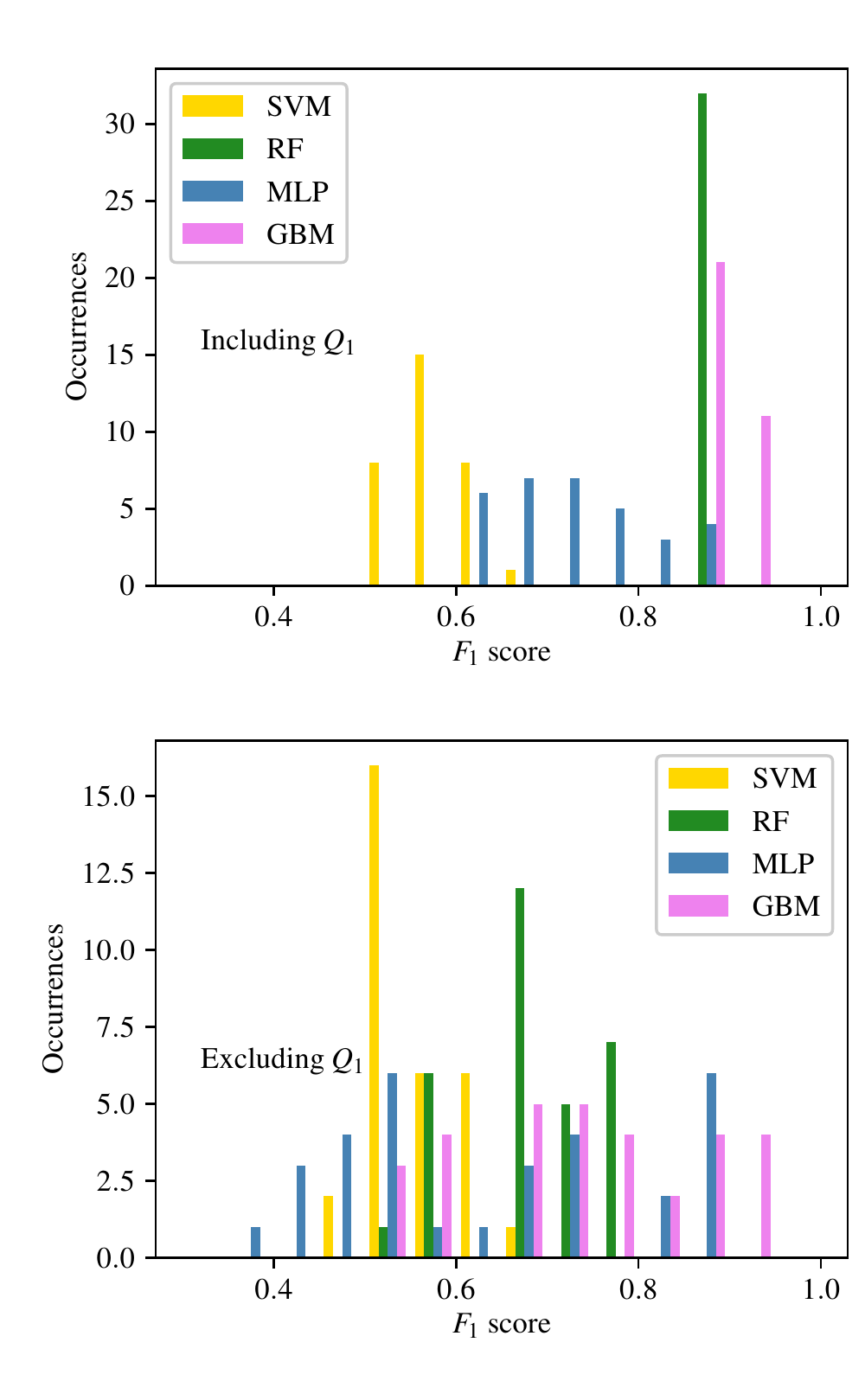}
\caption{The histograms of the $F_1$ scores achieved for CAS classification are shown for all input measurement combinations that include $Q_1$ in the upper plot, and exclude $Q_1$ in the lower plot.}
\label{figure_carotid_carotid}
\end{figure}

\begin{figure}[htbp]
\centering
\includegraphics[width=4in]{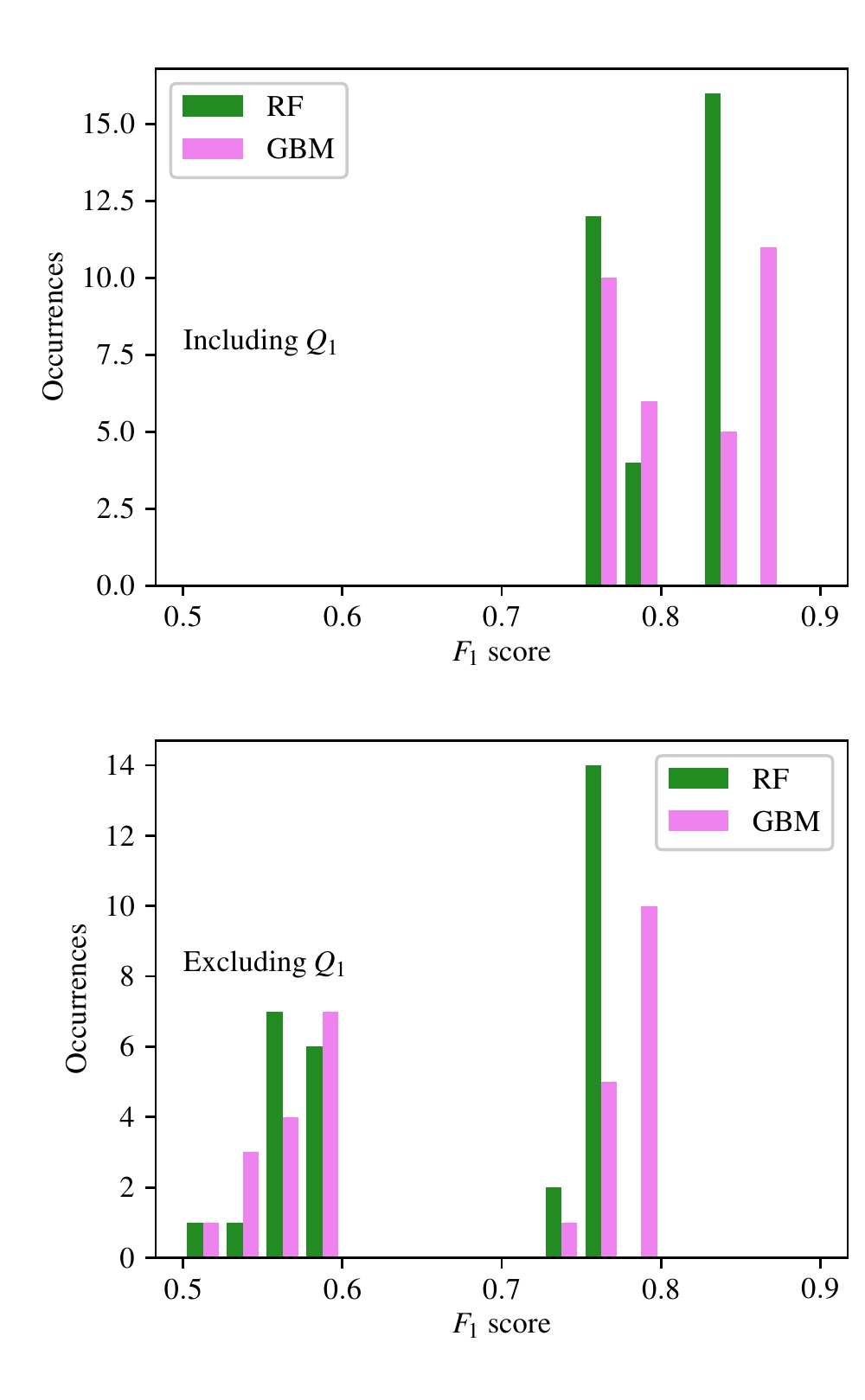}
\caption{The histograms of the $F_1$ scores achieved for SAS classification are shown for all input measurement combinations that include $Q_1$ in the upper plot, and exclude $Q_1$ in the lower plot.}
\label{figure_carotid_SS}
\end{figure}

\begin{figure}[htbp]
\centering
\includegraphics[width=4in]{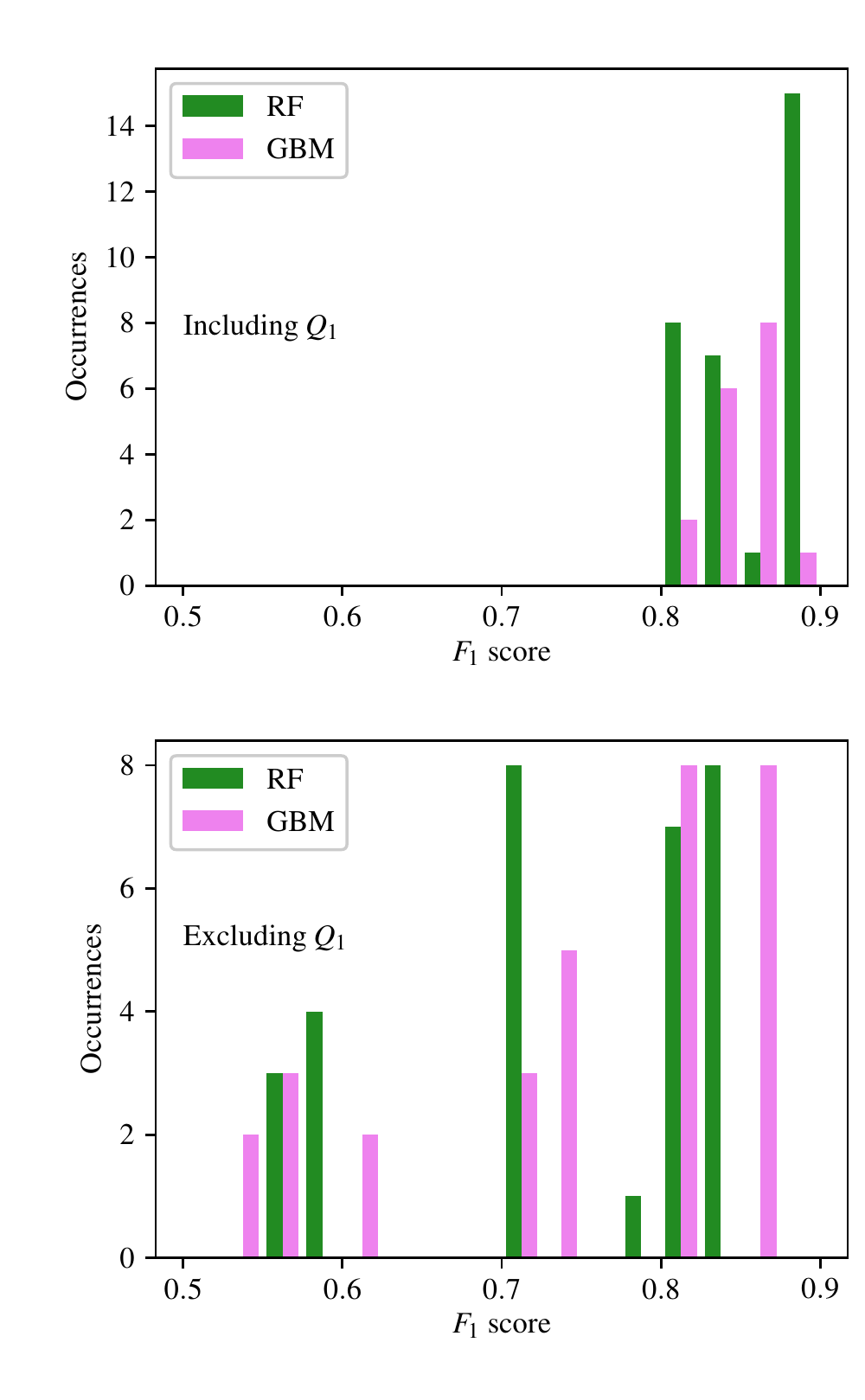}
\caption{The histograms of the $F_1$ scores achieved for PAD classification are shown for all input measurement combinations that include $Q_1$ in the upper plot, and exclude $Q_1$ in the lower plot.}
\label{figure_carotid_PAD}
\end{figure}

\begin{figure}[htbp]
\centering
\includegraphics[width=4in]{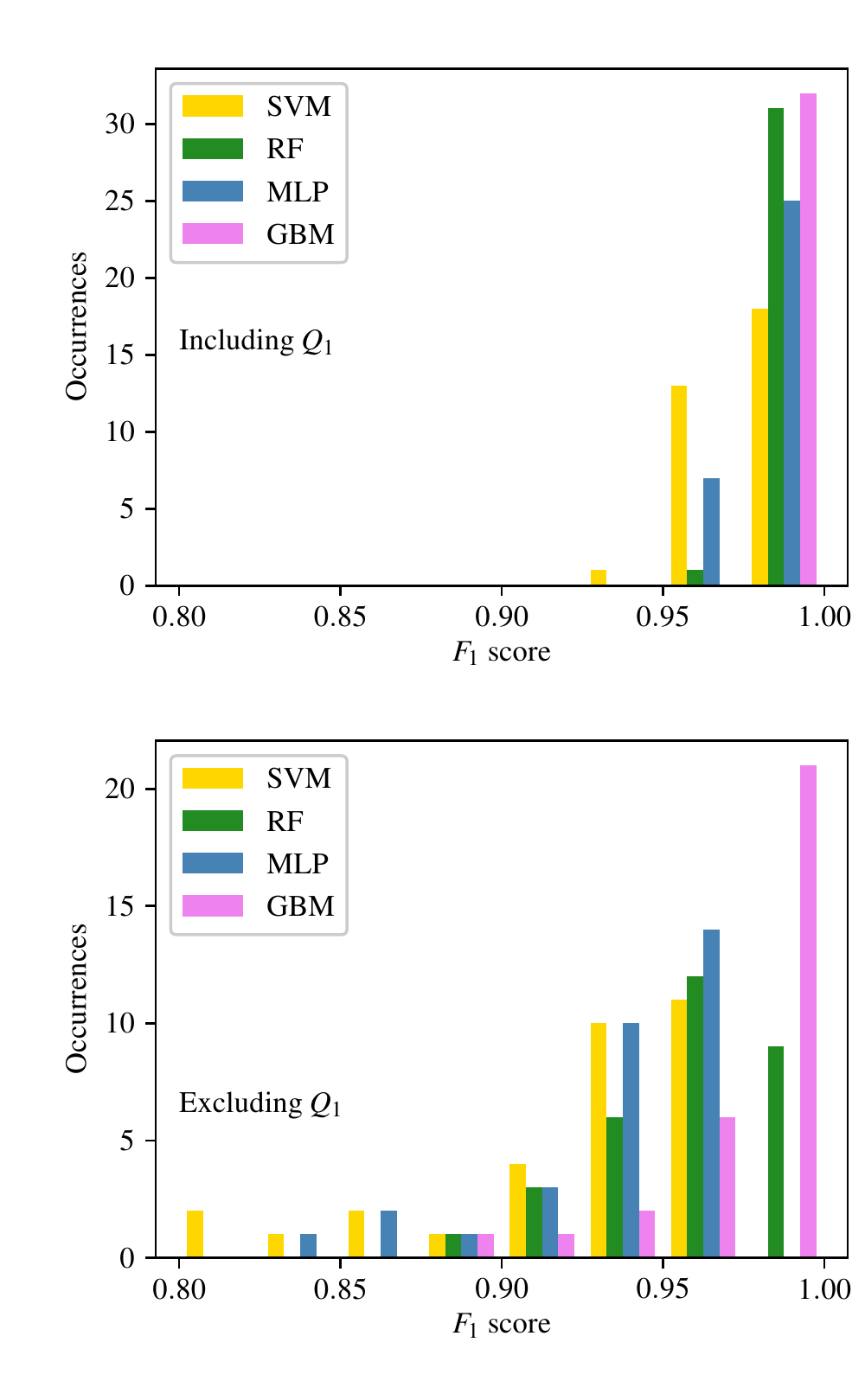}
\caption{The histograms of the $F_1$ scores achieved for AAA classification are shown for all input measurement combinations that include $Q_1$ in the upper plot, and exclude $Q_1$ in the lower plot.}
\label{figure_carotid_AAA}
\end{figure}

Appendices \ref{appendix_CS}--\ref{appendix_AAA}, along with the above analysis show that classifiers trained using flow-rates in the common carotid arteries ($Q_1$) consistently produce the highest accuracy. To analyse this further, the $F_1$ scores of classifiers with combinations that include and exclude $Q_1$ are separated and compared for CAS, SAS, PAD, and AAA in Figures \ref{figure_carotid_carotid}, \ref{figure_carotid_SS}, \ref{figure_carotid_PAD}, and \ref{figure_carotid_AAA} respectively. These figures show the the histograms of the $F_1$ scores, i.e. the number of occurrences/classifiers/combinations including and excluding $Q_1$ against $F_1$ score buckets. For each disease form, results are only shown for the classification methods that consistently produce good results for the corresponding disease form. The figures show a clear positive shift in the histograms when $Q_1$ is included, pointing to the particularly informative nature of $Q_1$. Other behaviours observed from these figures are: 
\begin{itemize}[leftmargin=*]
\item While there is generally an increase in $F_1$ score when including $Q_1$, it is also simultaneously observed that the maximum accuracies are relatively less sensitive to the inclusion of $Q_1$.
\item The greatest distinction between $F_1$ scores  when  including or excluding $Q_1$ is seen for CAS classification when using the RF method. There is no overlap between the two RF histograms in Figure \ref{figure_carotid_carotid}.
\item Observing the lower plots in Figures \ref{figure_carotid_SS} and \ref{figure_carotid_PAD}, a clear subgroup of low-accuracy classifiers can be seen when excluding $Q_1$ for SAS and PAD, which does not exist when including $Q_1$.
\end{itemize}

\subsection{Feature importance}
\label{sec_FI}
An important aspect of the GB method is that the measurement importance,  which determines the influence that individual measurements have towards classification, can be computed. This split-improvement feature importance \cite{zhou2019unbiased} of a feature can be thought of as the contribution of that feature to the total information gain achieved in a decision tree, averaged across all the trees in the ensemble. A high feature importance suggests that the given feature is contributing heavily to the classification accuracies achieved. As the features provided to the GB classifiers are the FS coefficients describing the haemodynamic profiles, the total importance of each bilateral pressure or flow-rate measurement is found by summing the feature importance of the associated 22 FS coefficients. The total importance of each input measurement for each disease form is shown in Table \ref{table_FI}.
\begin{table}[htbp]
\begin{center}
\def\arraystretch{1.2}
\begin{tabular}{ |c | c c c c c c |}
\hline
 & \textbf{$\bm{Q}_1$ (\%)} & \textbf{$\bm{Q}_2$ (\%)} & \textbf{$\bm{Q}_3$ (\%)} &  \textbf{$\bm{P}_1$ (\%)} & \textbf{$\bm{P}_2$ (\%)} & \textbf{$\bm{P}_3$ (\%)}\\
\hline
\textbf{CAS} & 67.38 & 8.02 & 3.89 & 11.07 & 7.692 & 1.93\\
\textbf{SAS} & 41.90 & 29.98 & 8.40 & 6.80 & 5.97 & 6.921\\
\textbf{PAD} & 38.01 & 15.98 & 31.11 & 4.62 & 4.63 & 5.62\\
\textbf{AAA} & 69.34 & 19.10 & 4.95 & 2.41 & 2.61 & 1.55\\
\hline 
\end{tabular}
\caption{The total importance of each input measurement, based on the GB classifiers provided with all six measurements.}
\label{table_FI}
\end{center}
\end{table} 
Three important observations  from this table are: 
\begin{itemize}[leftmargin=*]
\item The input measurement $Q_1$ consistently produces the highest importance for all forms of disease. This finding supports the findings of Section \ref{section_carotid_flow}.
\item The importance of each input measurement changes between disease forms based on the spatial proximity to the disease location. Generally, the measurements in close proximity to the disease location have higher importance. For example the importance of $Q_3$ (flow-rate in the femoral arteries) is highest for PAD classification (see Figure \ref{fig_full_network} for locations of disease and measurements). Similarly, $P_1$ (pressure in carotid arteries) has highest importance for CAS and SAS.
\item The feature importances, when viewed in collection, also shed some light on why $Q_1$ is important for SAS and PAD even though the measurement location is far from the disease location. For SAS, the two most informative measurements are $Q_1$ and $Q_2$, and for PAD these are $Q_1$ and $Q_3$. From Figure \ref{fig_full_network}, it is clear that these combinations form pairs of flow-rates before and after/at the disease location. Thus the measurement locations bound the disease location to provide more information on the presence of disease.

\end{itemize}

\subsection{Lower severity aneurysms}
\label{sec_low_sev_AAA}

\begin{sidewaysfigure*}
\centering
\vspace{1cm}
\includegraphics[width=9in]{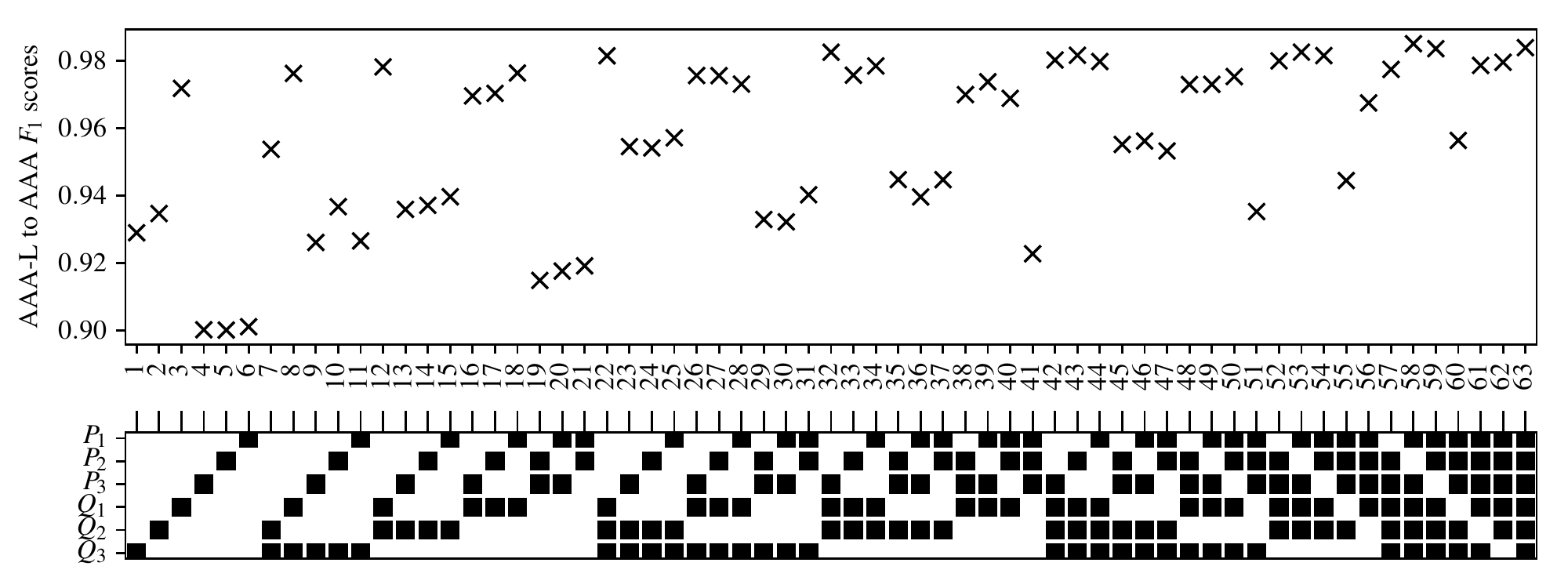}
\caption{The ratios of the $F_1$ scores for AAA-L classification relative to AAA classification, when providing each combination of input measurements are shown. Measurements included within each combination are highlighted with a black square.}
\label{fig_combination_ratios}
\end{sidewaysfigure*}

In Section \ref{sec_AAA_num_inputs} it is found that AAAs can be classified to a very high levels of accuracy with only one input measurement. Whether these accuracies are affected when lower severity aneurysms are considered is assessed here.
For this assessment, a new lower severity AAA VPD, referred to as $\text{VPD}_{\text{AAA-L}}$, is created in an  identical manner to the other diseased databases (see section \ref{sec_unhealthy_VPD}), with the following two differences:
\begin{itemize}[leftmargin=*]
\item The severity of aneurysms introduced into the virtual subjects (see Section \ref{section_disease_parameterisation}) is sampled from a uniform distribution bounded as follows: $3.0 \leq \mathcal{S}_{\text{aneurysm}} \leq 7.0$.
\item To reduce the computational expense associated with the creation of virtual patients, the size of $\text{VPD}_{\text{AAA-L}}$ is restricted to 5,000 VPs. 
\end{itemize}
 A combination search is carried out with only the GB method as it is the best overall method. 
The $F_1$ scores, sensitivities, and specificities achieved by all the measurement combinations are presented in Appendix \ref{appendix_LSA}. For comparison, the GB $F_1$ scores for all forms of disease  (including AAA-L) are shown in Appendix \ref{appendix_GB_combo}. The ratios of the GB $F_1$ scores achieved for AAA-L classification relative to AAA classification are shown in Figure \ref{fig_combination_ratios}.
 The observations from this figure are: 
 \begin{itemize}[leftmargin=*]
 \item The $F_1$ scores for AAA-L classification are consistently lower (ranging from 1\% to 10\% lower) than that for AAA classification. This finding supports the physiological expectation that the significance of biomarkers in pressure and flow-rate profiles is proportion to the severity.   
 
 \item The ratios of $F_1$ scores are lowest for combinations of inputs that predominantly rely on pressure measurements. This suggests that pressure measurements are, in general, less informative about disease severity. This is in support of the, generally, lower feature importance of pressure measurements in Table \ref{table_FI}.
 
 \item The $F_1$ score ratios are highest for input combinations that include $Q_1$. This finding further suggests that $Q_1$ contains consistent biomarkers.
 \item The ratios range between 0.9 and 0.99, implying a maximum degradation of only 10\% relative to high-severity classification accuracies. Thus, even in the low-severity aneurysms many combinations of classifiers achieve $F_1$ scores higher than 0.95 and corresponding sensitivities and specificities larger than 95\%.
 \end{itemize}

\subsection{Unilateral aneurysm measurement tests}

Hitherto, all ML classifiers used bilateral measurements, \textit{i.e.} both the right and left instances of each measurement were simultaneously provided. Here, the ability of unilateral measurements, \textit{i.e.} only the right or left instance of a measurement, to detect AAAs is assessed. This analysis is restricted to the GB method as it consistently outperforms other methods.
 GB classifiers are trained and tested to detect AAAs using four different unilateral measurements:
\begin{itemize}[leftmargin=*]
\item \textbf{Flow-rate in the right carotid artery}, shown in Figure \ref{fig_full_network} as $Q_1^{\text{(R)}}$.
\item \textbf{Flow-rate in the left carotid artery}, shown in Figure \ref{fig_full_network} as $Q_1^{\text{(L)}}$.
\item \textbf{Pressure in the right radial artery}, shown in Figure \ref{fig_full_network} as $P_3^{\text{(R)}}$.
\item \textbf{Pressure in the left radial artery}, shown in Figure \ref{fig_full_network} as $P_3^{\text{(L)}}$.
\end{itemize}
Carotid artery flow-rate is chosen as it has been shown to be the best measurement for disease classification. 
Radial artery pressure is chosen due to the location of the radial artery on the human wrist. While wearable devices capable of measuring continuous radial pressure profiles do not  currently exist (to the authors' knowledge), if AAAs can be detected to satisfactory accuracies using these measurements, it may suggest the possibility of future home monitoring of abdominal aortic health through the development of such wearables. The sensitivities and specificities achieved 
by the four unilateral GB classifiers are shown in Table \ref{table_unilaterial_results}.
It shows that relative to the bilateral case, while there is a decrease in the classification accuracies, the magnitude of the decrease is less than 10\%. 
This finding suggests that there may be sufficient biomarkers of AAA presence captured within the intra-measurement details of a single pressure or flow-rate profile. The fact that similar accuracies  are achieved with either the right or left instances of any measurement is likely due to physiological symmetry. While there are some minor asymmetries between the right and left upper extremities, due to the topology of the arterial network (as shown in Figure \ref{fig_full_network}) changes to the cross sectional area of the abdominal aorta are expected to produce relatively consistent changes in both the right and left side of the body. 

%
\begin{table}[htbp]
\begin{center}
\def\arraystretch{1.2}
\begin{tabular}{| c | c | c c |}
\hline
& \textbf{Side} & \textbf{Sensitivity} & \textbf{Specificity}\\
\hline
\textbf{Carotid} & Right & 0.9369 & 0.9161 \\
\textbf{flow-rate} & Left & 0.9065 & 0.9146 \\
($Q_1$) & Both & 0.9799 & 0.9811 \\
 \hline
\textbf{Radial} & Right & 0.8356 & 0.8533 \\
\textbf{pressure} & Left & 0.8633 & 0.8605 \\
($P_3$) & Both & 0.9202 & 0.9248\\
\hline 
\end{tabular}
\caption{The sensitivities and specificities achieved when using the measurements of flow-rate in the right, left, and both CAs and pressure in the right, left, and both radial arteries.}
\label{table_unilaterial_results}
\end{center}
\end{table}

\section{Conclusions}

The main conclusion of this study is that machine learning methods are suitable for detection of arterial disease---both stenoses and aneurysms---from peripheral measurements of pressure and flow-rates across the network. Amongst various ML methods, it is found that tree-based methods of Random Forest and Gradient Boosting perform best for this application. Across the different forms of disease, the Gradient Boosting method outperforms Random Forest, Support Vector Machine, Naive Bayes, Logistic Regression, and even the deep learning method of Multi-layer Perceptron.

It is demonstrated that maximum $F_1$ scores larger than 0.9 are achievable for CAS and PAD, larger than 0.85 for SAS, and larger than 0.98 for both low- and high-severity AAAs. The corresponding sensitivities and specificities are also both larger than 90\% for CAS and PAD, larger than 85\% for SAS, and larger than 98\% for both low- and high-severity AAAs. 
%
%
While these maximum scores are for the case when all the six measurements are used, it is also shown that the performance degradation is less than 5\% when using only three measurements and less than 10\% when using only two measurements, as long as the these measurements are carefully chosen in specific combinations. For the case of AAA, it is further demonstrated that when only a single measurement (either on the left or right side) is used, $F_1$ scores larger than 0.85 and corresponding sensitivities and specificities larger than 85\% are achievable. This aspect encourages the application of AAA monitoring and/or screening through the use of a wearable device. Finally, it is shown through the analysis of several classifiers and feature-importance that, amongst the measurements, the carotid artery flow-rate is a particularly informative measurement to detect the presence of all the four forms of disease considered.

\section{Limitations \& future work}
While the results are encouraging, they are produced on a virtual cohort of subjects. Even though the database is physiologically realistic and carefully constructed, it may be that real patient behaviour differs from those in the VPD. Therefore, future steps should be in applying the trained classifiers here directly to a small cohort of real-patient measurements. The effect of measurement errors and biases is ignored in this study. This aspect can also be considered in future studies, along with relaxing the assumption of mutually exclusive disease. Thus, using either real or virtual databases, classifiers should be built to detect not only the presence of disease, but also identify the type of disease (potentially concomitant disease in multiple locations), its location, and its severity.  Further improvements can be also made by further optimising the architectures of the machine and deep learning methods to aim for higher accuracies with fewer, potentially noise- and bias-corrupted, measurements.

\section*{Funding}
This work is supported by an EPSRC studentship ref. EP/N509553/1 and an EPSRC grant ref. EP/R010811/1.

\printbibliography

\clearpage

\appendix
\footnotesize

\section{CAS combination search results}
\label{appendix_CS}

The $F_1$ scores, sensitivities, and specificities achieved for CAS classification when using each of the six ML methods are shown in Table \ref{table_F1_stenosis_CS}, \ref{table_healthy_stenosis_CS}, and \ref{table_unhealthy_stenosis_CS} respectively.

\begin{longtable}{|c|c c c c c c|}

\hline
& \multicolumn{6}{c|}{\textbf{Classification method}}\\
\textbf{Input combination} & \textbf{NB} & \textbf{LR} & \textbf{SVM} & \textbf{RF} & \textbf{MLP} & \textbf{GB}\\
\hline
\endhead
\hline
\endfoot
\endlastfoot
 $Q_3$ & 0.5547 & 0.5110 & 0.5157 & 0.5807 & 0.4365 & 0.5606 \\
 $Q_2$ & 0.5105 & 0.5080 & 0.4955 & 0.6858 & 0.4410 & 0.6565 \\
 $Q_1$ & 0.5676 & 0.5033 & 0.5953 & 0.8809 & 0.6459 & 0.8521 \\
 $P_3$ & 0.4927 & 0.5023 & 0.4991 & 0.5441 & 0.4805 & 0.5131 \\
 $P_2$ & 0.4413 & 0.5066 & 0.5260 & 0.5628 & 0.3741 & 0.5412 \\
 $P_1$ & 0.5473 & 0.4917 & 0.5712 & 0.6681 & 0.7013 & 0.7082 \\
 $Q_3$, $Q_2$ & 0.5684 & 0.4955 & 0.5104 &  0.6955 & 0.4915 & 0.6889 \\
 $Q_3$, $Q_1$ & 0.4831 & 0.5050 & 0.5544 &  0.8790 & 0.6944 & 0.8629 \\
\end{longtable}

\clearpage

\begin{longtable}{|c|c c c c c c|}

\hline
& \multicolumn{6}{c|}{\textbf{Classification method}}\\
\textbf{Input combination} & \textbf{NB} & \textbf{LR} & \textbf{SVM} & \textbf{RF} & \textbf{MLP} & \textbf{GB}\\ 
\hline
\endhead
\hline
\endfoot
\endlastfoot
 $Q_3$ & 0.1531 & 0.5527 & 0.4283 & 0.5572
 & 0.6084 & 0.5736 \\
 $Q_2$ & 0.6641 & 0.5097 & 0.6418 & 0.6575 & 0.6228 & 0.6744 \\
 $Q_1$ & 0.5426 & 0.4525 & 0.5694 & 0.8704 & 0.4243 & 0.8547 \\
 $P_3$ & 0.5024 & 0.5098 & 0.4999 & 0.5139 & 0.5355 & 0.5158 \\
 $P_2$ & 0.6490 & 0.4979 & 0.5052 & 0.5366 & 0.7038 & 0.5491 \\
 $P_1$ & 0.2410 & 0.4992 & 0.6588 & 0.6510
 & 0.7052 & 0.7215 \\
 $Q_3$, $Q_2$ & 0.5055 & 0.5035 & 0.5651 &  0.6655 & 0.5439 & 0.7044 \\ 
 $Q_3$, $Q_1$ & 0.7217 & 0.5094 & 0.5461 &  0.8681 & 0.6288 & 0.8661 \\
\end{longtable}

\clearpage

\begin{longtable}{|c|c c c c c c|}

\hline
& \multicolumn{6}{c|}{\textbf{Classification method}}\\
\textbf{Input combination} & \textbf{NB} & \textbf{LR} & \textbf{SVM} & \textbf{RF} & \textbf{MLP} & \textbf{GB} \\
\hline
\endhead
\hline
\endfoot
\endlastfoot
 $Q_3$ & 0.7090 & 0.4968 & 0.5462 & 0.5904 & 0.3886 & 0.5556 \\
 $Q_2$ & 0.4579 & 0.5075 & 0.4474 & 0.7006 & 0.3896 & 0.6479 \\
 $Q_1$ & 0.5776 & 0.5204 & 0.6063 & 0.8893 & 0.7517 & 0.8502 \\
 $P_3$ & 0.4896 & 0.4998 & 0.4989 & 0.5555 & 0.4632 & 0.5122 \\
 $P_2$ & 0.3826 & 0.5096 & 0.5335 & 0.5732 & 0.2983 & 0.5384 \\
 $P_1$ & 0.6628 & 0.4893 & 0.5363 & 0.6767 & 0.6993 & 0.7010 \\
 $Q_3$, $Q_2$ & 0.5935 & 0.4929 & 0.4917 &  0.7116 & 0.4745 & 0.6808 \\
 $Q_3$, $Q_1$ & 0.4072 & 0.5036 & 0.5576 &  0.8876 & 0.7293 & 0.8605 \\
\end{longtable}

\section{SAS combination search results}
\label{appendix_SS}

The $F_1$ scores, sensitivities, and specificities achieved for SAS classification when using each of the six ML methods are shown in Table \ref{table_F1_stenosis_SS}, \ref{table_healthy_stenosis_SS}, and \ref{table_unhealthy_stenosis_SS} respectively.

\begin{longtable}{|c|c c c c c c|}

\hline
& \multicolumn{6}{c|}{\textbf{Classification method}}\\
\textbf{Input combination} & \textbf{NB} & \textbf{LR} & \textbf{SVM} & \textbf{RF} & \textbf{MLP} & \textbf{GB}\\
\hline
\endhead
\hline
\endfoot
\endlastfoot
 $Q_3$ & 0.5041 & 0.5288 & 0.4897 & 0.5723 & 0.5403 & 0.5592 \\
 $Q_2$ & 0.4681 & 0.5004 & 0.4839 & 0.7577 & 0.5691 & 0.7415 \\
 $Q_1$ & 0.3799 & 0.5028 & 0.4923 & 0.7779 & 0.6176 & 0.7529 \\
 $P_3$ & 0.4931 & 0.4972 & 0.5097 & 0.5530 & 0.5474 & 0.5331 \\
 $P_2$ & 0.4698 & 0.4990 & 0.5528 & 0.5627 & 0.4895 & 0.5453 \\
 $P_1$ & 0.5344 & 0.5023 & 0.5035 & 0.5171 & 0.5571 & 0.5060 \\
 $Q_3$, $Q_2$ & 0.4529 & 0.5136 & 0.5075 & 0.7623 & 0.4939 & 0.7608 \\
 $Q_3$, $Q_1$ & 0.4588 & 0.4893 & 0.5053 & 0.7814 & 0.5414 & 0.7758 \\
\end{longtable}

\clearpage

\begin{longtable}{|c|c c c c c c|}

\hline
& \multicolumn{6}{c|}{\textbf{Classification method}}\\
\textbf{Input combination} & \textbf{NB} & \textbf{LR} & \textbf{SVM} & \textbf{RF} & \textbf{MLP} & \textbf{GB}\\ 
\hline
\endhead
\hline
\endfoot
\endlastfoot
 $Q_3$ & 0.2997 & 0.4576 & 0.5129 & 0.5678 & 0.4059 & 0.5585 \\
 $Q_2$ & 0.5460 & 0.5348 & 0.6918 & 0.7517 & 0.3839 & 0.7366 \\
 $Q_1$ & 0.7074 & 0.4613 & 0.6338 & 0.7582 & 0.1873 & 0.7224 \\
 $P_3$ & 0.4402 & 0.5127 & 0.5616 & 0.5453 & 0.3978 & 0.5431 \\
 $P_2$ & 0.5140 & 0.4981 & 0.4783 & 0.5629 & 0.5704 & 0.5717 \\
 $P_1$ & 0.4446 & 0.4836 & 0.4741 & 0.5177 & 0.3803 & 0.5244 \\
 $Q_3$, $Q_2$ & 0.5683 & 0.4928 & 0.5411 &  0.7612 & 0.5901 & 0.7585 \\ 
 $Q_3$, $Q_1$ & 0.4887 & 0.4947 & 0.5036 &  0.7630 & 0.4709 & 0.7504 \\
\end{longtable}

\clearpage

\begin{longtable}{|c|c c c c c c|}

\hline
& \multicolumn{6}{c|}{\textbf{Classification method}}\\
\textbf{Input combination} & \textbf{NB} & \textbf{LR} & \textbf{SVM} & \textbf{RF} & \textbf{MLP} & \textbf{GB} \\
\hline
\endhead
\hline
\endfoot
\endlastfoot
 $Q_3$ & 0.5731 & 0.5544 & 0.4823 & 0.5742 & 0.5901 & 0.5596 \\
 $Q_2$ & 0.4444 & 0.4890 & 0.4176 & 0.7615 & 0.6429 & 0.7445 \\
 $Q_1$ & 0.3032 & 0.5168 & 0.4462 & 0.7905 & 0.8099 & 0.7714 \\
 $P_3$ & 0.5105 & 0.4921 & 0.4920 & 0.5560 & 0.6038 & 0.5295 \\
 $P_2$ & 0.4563 & 0.4993 & 0.5814 & 0.5627 & 0.4633 & 0.5355 \\
 $P_1$ & 0.5672 & 0.5087 & 0.5135 & 0.5170
& 0.6254 & 0.4999 \\
 $Q_3$, $Q_2$ & 0.4192 & 0.5208 & 0.4962 &  0.7631 & 0.4624 & 0.7623 \\
 $Q_3$, $Q_1$ & 0.4499 & 0.4876 & 0.5059 & 0.7932 & 0.5676 & 0.7920 \\
\end{longtable}

\section{PAD combination search results}
\label{appendix_PAD}

The $F_1$ scores, sensitivities, and specificities achieved for PAD classification when using each of the six ML methods are shown in Table \ref{table_F1_stenosis_PAD}, \ref{table_healthy_stenosis_PAD}, and \ref{table_unhealthy_stenosis_PAD} respectively.

\begin{longtable}{|c|c c c c c c|}

\hline
& \multicolumn{6}{c|}{\textbf{Classification method}}\\
\textbf{Input combination} & \textbf{NB} & \textbf{LR} & \textbf{SVM} & \textbf{RF} & \textbf{MLP} & \textbf{GB}\\
\hline
\endhead
\hline
\endfoot
\endlastfoot
 $Q_3$ & 0.5017 & 0.5115 & 0.6645 & 0.8224 & 0.6897 & 0.8169 \\
 $Q_2$ & 0.5621 & 0.5222 & 0.5266 & 0.7127 & 0.4734 & 0.7076 \\
 $Q_1$ & 0.3927 & 0.4822 & 0.5310 & 0.8240 & 0.4713 & 0.8183 \\
 $P_3$ & 0.5162 & 0.5053 & 0.5182 & 0.5613 & 0.4131 & 0.5406 \\
 $P_2$ & 0.5030 & 0.4954 & 0.5242 & 0.5753 & 0.4741 & 0.5529 \\
 $P_1$ & 0.4290 & 0.5031 & 0.5038 & 0.5517 & 0.5487 & 0.5335 \\
 $Q_3$, $Q_2$ & 0.4740 & 0.5099 & 0.5926 & 0.8480 & 0.7040 & 0.8557 \\
 $Q_3$, $Q_1$ & 0.5355 & 0.4965 & 0.5786 & 0.8959 & 0.7254 & 0.9041 \\
\end{longtable}

\clearpage

\begin{longtable}{|c|c c c c c c|}

\hline
& \multicolumn{6}{c|}{\textbf{Classification method}}\\
\textbf{Input combination} & \textbf{NB} & \textbf{LR} & \textbf{SVM} & \textbf{RF} & \textbf{MLP} & \textbf{GB}\\ 
\hline
\endhead
\hline
\endfoot
\endlastfoot
 $Q_3$ & 0.3598 & 0.5048 & 0.6806 & 0.8219 & 0.5998 & 0.8188 \\
 $Q_2$ & 0.5441 & 0.4878 & 0.5879 & 0.6858 & 0.5536 & 0.6922 \\
 $Q_1$ & 0.5735 & 0.5026 & 0.6065 & 0.8126 & 0.5959 & 0.8140 \\
 $P_3$ & 0.4246 & 0.4935 & 0.5472 & 0.5358 & 0.6388 & 0.5425 \\
 $P_2$ & 0.4565 & 0.4985 & 0.5368 & 0.5532 & 0.5572 & 0.5576 \\
 $P_1$ & 0.6253 & 0.5001 & 0.5571 & 0.5245 & 0.3899 & 0.5261 \\
 $Q_3$, $Q_2$ & 0.5595 & 0.4912 & 0.6297 &  0.8414 & 0.7176 & 0.8532 \\ 
 $Q_3$, $Q_1$ & 0.4753 & 0.5087 & 0.6324 &  0.8825 & 0.7460 & 0.8950 \\
\end{longtable}

\clearpage

\begin{longtable}{|c|c c c c c c|}

\hline
& \multicolumn{6}{c|}{\textbf{Classification method}}\\
\textbf{Input combination} & \textbf{NB} & \textbf{LR} & \textbf{SVM} & \textbf{RF} & \textbf{MLP} & \textbf{GB} \\
\hline
\endhead
\hline
\endfoot
\endlastfoot
 $Q_3$ & 0.5493 & 0.5139 & 0.6566 & 0.8228 & 0.7371 & 0.8157 \\
 $Q_2$ & 0.5692 & 0.5344 & 0.5047 & 0.7276 & 0.4486 & 0.7161 \\
 $Q_1$ & 0.3486 & 0.4758 & 0.5038 & 0.8320 & 0.4329 & 0.8214 \\
 $P_3$ & 0.5481 & 0.5093 & 0.5081 & 0.5713 & 0.3544 & 0.5399 \\
 $P_2$ & 0.5187 & 0.4945 & 0.5198 & 0.5843 & 0.4484 & 0.5512 \\
 $P_1$ & 0.3754 & 0.5042 & 0.4859 & 0.5622 & 0.6088 & 0.5362 \\
 $Q_3$, $Q_2$ & 0.4475 & 0.5164 & 0.5770 &  0.8529 & 0.6967 & 0.8576 \\
 $Q_3$, $Q_1$ & 0.5576 & 0.4926 & 0.5568 &  0.9068 & 0.7137 & 0.9117 \\
\end{longtable}

\section{AAA combination search results}
\label{appendix_AAA}

The $F_1$ scores, sensitivities, and specificities achieved for AAA classification when using each of the six ML methods are shown in Table \ref{table_F1_stenosis_AAA}, \ref{table_healthy_stenosis_AAA}, and \ref{table_unhealthy_stenosis_AAA} respectively.

\begin{longtable}{|c|c c c c c c|}

\hline
& \multicolumn{6}{c|}{\textbf{Classification method}}\\
\textbf{Input combination} & \textbf{NB} & \textbf{LR} & \textbf{SVM} & \textbf{RF} & \textbf{MLP} & \textbf{GB}\\
\hline
\endhead
\hline
\endfoot
\endlastfoot
 $Q_3$ & 0.4670 & 0.4881 & 0.8454 & 0.9095 & 0.8606 & 0.9294 \\
 $Q_2$ & 0.5754 & 0.4952 & 0.8246 & 0.9516 & 0.9092 & 0.9640 \\
 $Q_1$ & 0.4440 & 0.4843 & 0.9481 & 0.9741 & 0.9697 & 0.9805 \\
 $P_3$ & 0.4999 & 0.5102 & 0.8664 & 0.9027 & 0.8692 & 0.9226 \\
 $P_2$ & 0.5782 & 0.4944 & 0.8717 & 0.9087 & 0.8793 & 0.9311 \\
 $P_1$ & 0.4790 & 0.4826 & 0.8212 & 0.8771 & 0.8416 & 0.8884 \\
 $Q_3$, $Q_2$ & 0.3850 & 0.4983 & 0.8895 & 0.9753 & 0.9249 & 0.9843 \\
 $Q_3$, $Q_1$ & 0.4982 & 0.5029 & 0.9521 & 0.9840 & 0.9749 & 0.9919 \\
\end{longtable}

\clearpage

\begin{longtable}{|c|c c c c c c|}

\hline
& \multicolumn{6}{c|}{\textbf{Classification method}}\\
\textbf{Input combination} & \textbf{NB} & \textbf{LR} & \textbf{SVM} & \textbf{RF} & \textbf{MLP} & \textbf{GB}\\ 
\hline
\endhead
\hline
\endfoot
\endlastfoot
 $Q_3$ & 0.5683 & 0.5120 & 0.8568 & 0.8878 & 0.8661 & 0.9300 \\
 $Q_2$ & 0.5738 & 0.5089 & 0.8136 & 0.9355 & 0.9100 & 0.9638 \\
 $Q_1$ & 0.4451 & 0.4962 & 0.9517 & 0.9654 & 0.9673 & 0.9799 \\
 $P_3$ & 0.4846 & 0.5035 & 0.8785 & 0.8765 & 0.8660 & 0.9202 \\
 $P_2$ & 0.4451 & 0.5110 & 0.8712 & 0.9005 & 0.8818 & 0.9352 \\
 $P_1$ & 0.6616 & 0.4902 & 0.8491 & 0.8514 & 0.8308 & 0.8770 \\
 $Q_3$, $Q_2$ & 0.5927 & 0.4676 & 0.8868 & 0.9652 & 0.9308 & 0.9835 \\ 
 $Q_3$, $Q_1$ & 0.5541 & 0.5333 & 0.9508 &  0.9757 & 0.9747 & 0.9907 \\
\end{longtable}

\clearpage

\begin{longtable}{|c|c c c c c c|}

\hline
& \multicolumn{6}{c|}{\textbf{Classification method}}\\
\textbf{Input combination} & \textbf{NB} & \textbf{LR} & \textbf{SVM} & \textbf{RF} & \textbf{MLP} & \textbf{GB} \\
\hline
\endhead
\hline
\endfoot
\endlastfoot
 $Q_3$ & 0.4362 & 0.4805 & 0.8371 & 0.9276 & 0.8565 & 0.9290 \\
 $Q_2$ & 0.5761 & 0.4908 & 0.8324 & 0.9663 & 0.9087 & 0.9643 \\
 $Q_1$ & 0.4437 & 0.4806 & 0.9450 & 0.9825 & 0.9720 & 0.9811 \\
 $P_3$ & 0.5050 & 0.5126 & 0.8572 & 0.9244 & 0.8718 & 0.9248 \\
 $P_2$ & 0.6324 & 0.4890 & 0.8722 & 0.9156 & 0.8775 & 0.9277 \\
 $P_1$ & 0.4215 & 0.4803 & 0.8018 & 0.8972 & 0.8496 & 0.8976 \\
 $Q_3$, $Q_2$ & 0.3355 & 0.5086 & 0.8917 & 0.9851 & 0.9200 & 0.9851 \\
 $Q_3$, $Q_1$ & 0.4797 & 0.4927 & 0.9533 & 0.9922 & 0.9751 & 0.9931 \\
\end{longtable}

\clearpage
\section{AAA-L combination search results}
\label{appendix_LSA}

The $F_1$ scores, sensitivities, and specificities achieved for AAA-L classification when employing the GB method are shown in Table \ref{table_F1_AAA_L}.

\begin{longtable}{|c|c c c|}

\hline
\textbf{Input combination} & \textbf{$F_1$} & \textbf{Sen.} & \textbf{Spec.} \\
\hline
\endhead
\hline
\endfoot
\endlastfoot
 $Q_3$  & 0.8633 & 0.8561 & 0.8689 \\
 $Q_2$  & 0.9010 & 0.9103 & 0.8934 \\
 $Q_1$  & 0.9528 & 0.9630 & 0.9436\\
 $P_3$  & 0.8305 & 0.8383 & 0.8250\\
 $P_2$  & 0.8380 & 0.8529 & 0.8274\\
 $P_1$  & 0.8005 & 0.7700 & 0.8209\\
 $Q_3$, $Q_2$  & 0.9387 & 0.9390 & 0.9385\\
 $Q_3$, $Q_1$  & 0.9683 & 0.9681 & 0.9685\\
\end{longtable}
\clearpage

\section{GB results for all disease forms}
\label{appendix_GB_combo}

\begin{figure}
\centering
\includegraphics[width=7in]{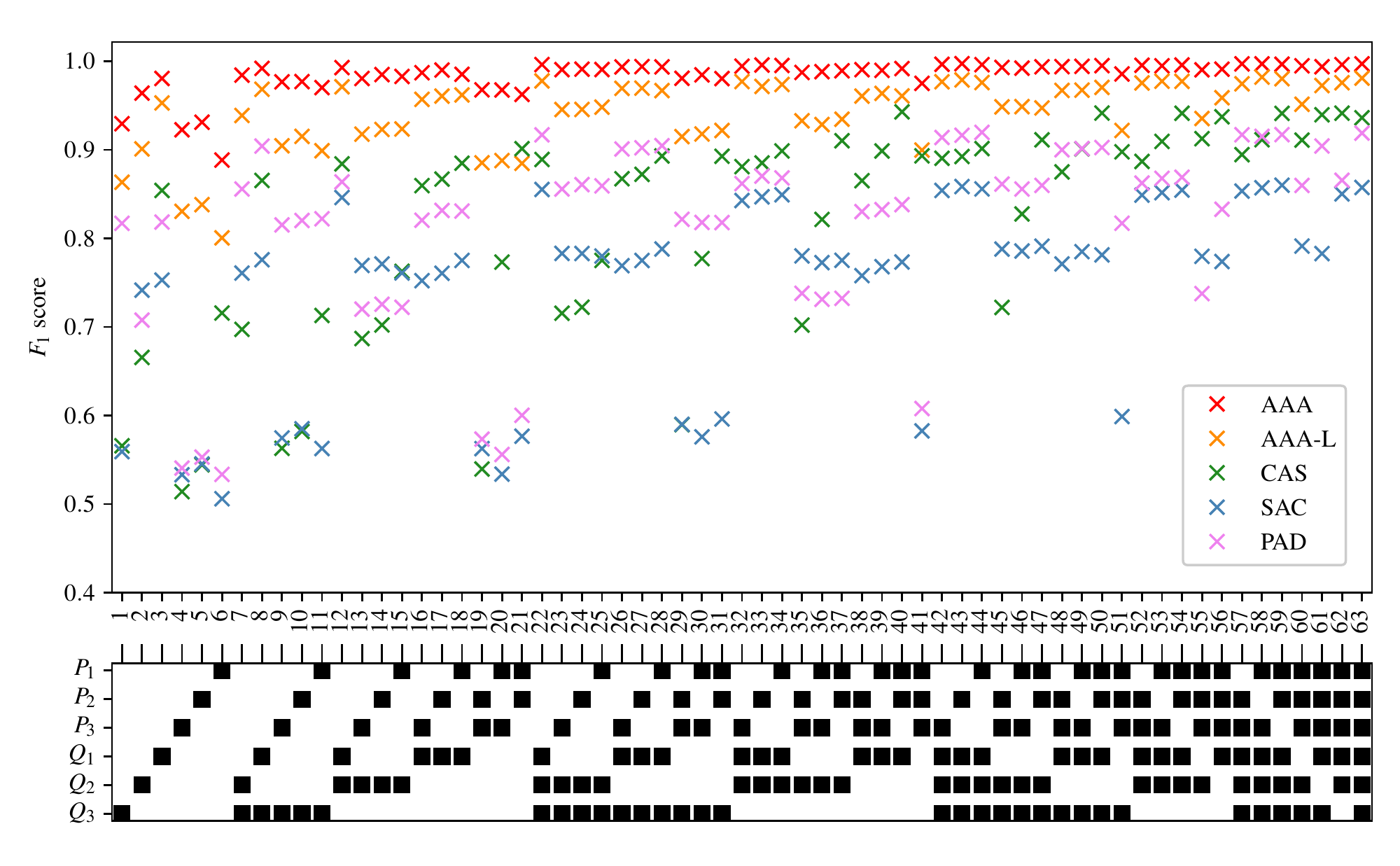}
\caption{The $F_1$ scores achieved for all disease forms when employing the GB method. Measurements included within each combination are highlighted with a black square.}
\label{fig_combination_inputs_AAA_high_low}
\end{figure}

The $F_1$ scores achieved for all forms of disease classification (including AAA-L) when providing each combination of input measurements are shown when employing the GB method in Figure \ref{fig_combination_inputs_AAA_high_low}.

\end{document}